%% file: main.tex
\documentclass[journal]{IEEEtran}

\usepackage{enumitem}
\usepackage[utf8]{inputenc} % allow utf-8 input
\usepackage[T1]{fontenc}    % use 8-bit T1 fonts
\usepackage{hyperref}       % hyperlinks
\usepackage{url}            % simple URL typesetting
\usepackage{graphicx, subfigure, float}
\usepackage{epstopdf, color}
\usepackage{booktabs,bm}       % professional-quality tables
\usepackage{amsfonts}       % blackboard math symbols
\usepackage{nicefrac}       % compact symbols for 1/2, etc.
\usepackage{microtype}      % microtypography
\usepackage[cmex10]{amsmath}
\usepackage{algorithm}
\usepackage{algpseudocode}
\usepackage{amssymb}
\usepackage{amsthm}
\usepackage{mathtools}
\usepackage{enumerate}
\usepackage{mathrsfs} %TO get \mathscr{X} for the state space
\usepackage{wrapfig}
\usepackage{makecell}
\usepackage{multirow}
\usepackage{multicol}
\usepackage{textcomp}

\title{Improving Actor-Critic Reinforcement Learning via Hamiltonian Monte Carlo Method}

\author{
\vspace*{10pt}
\IEEEauthorblockN{Duo Xu, \and Faramarz Fekri} \\
\IEEEauthorblockA{ 
School of Electrical and Computer Engineering\\
Georgia Institute of Technology\\
 Atlanta, GA, 30332
}
\vspace*{-10pt}
}

\begin{document}

\maketitle

\begin{abstract}
The actor-critic RL is widely used in various robotic control tasks. By viewing the actor-critic RL from the perspective of variational inference (VI), the policy network is trained to obtain the approximate posterior of actions given the optimality criteria.  
However, in practice, the actor-critic RL may yield suboptimal policy estimates due to the amortization gap and insufficient exploration. In this work, inspired by the previous use of Hamiltonian Monte Carlo (HMC) in VI, we propose to integrate the policy network of actor-critic RL with HMC, which is termed as {\it Hamiltonian Policy}. As such we propose to evolve actions from the base policy according to HMC, and our proposed method has many benefits. First, HMC can improve the policy distribution to better approximate the posterior and hence reduce the amortization gap. Second, HMC can also guide the exploration more to the regions of action spaces with higher Q values, enhancing the exploration efficiency. Further, instead of directly applying HMC into RL, we propose a new leapfrog operator to simulate the Hamiltonian dynamics. Finally, in safe RL problems, we find that the proposed method can not only improve the achieved return, but also reduce safety constraint violations by discarding potentially unsafe actions. 
With comprehensive empirical experiments on continuous control baselines, including MuJoCo and PyBullet Roboschool, we show that the proposed approach is a data-efficient and easy-to-implement improvement over previous actor-critic methods. 
\end{abstract}
%Codes are available at \url{https://drive.google.com/drive/folders/1ae3j_UMj2OwSlxGWp1_wKB8EMBV1ZBKE?usp=sharing}.

\input{part1}
\input{part2}

\input{part3}

\bibliographystyle{plain}
\bibliography{main}

%\EOD

\end{document}

%% file: part1.tex
\section{Introduction}
%Reinforcement learning (RL) algorithms involve policy evaluation and policy optimization \cite{suttonbartorlbook}. %Given a policy, one can estimate the value for each state or state-action pair by directly applying that policy, and given a value estimate, one can improve the policy by maximizing the value. This latter procedure, 
In continuous control, actor-critic RL algorithms are widely used in solving practical problems. However, searching optimal policies can be challenging due to instability and poor asymptotic performance. 
Specifically, most previous methods of actor-critic RL, such as KL regularization \cite{schulman2015trust,schulman2017proximal} and maximum policy entropy \cite{mnih2016asynchronous,fox2016taming}, essentially solve RL in the framework of variational inference (VI) \cite{levine2018reinforcement}, which infers a policy that yields high expected return while satisfying prior policy constraints. 
However, from this perspective, %when used with entropy or KL regularization, 
the policy network essentially performs amortized optimization \cite{gershman2014amortized,levine2018reinforcement}. It means that most actor-critic RL algorithms, such as soft actor-critic (SAC) \cite{haarnoja2018soft}, optimize a network to directly output the parameters of policy distribution which approximate the posterior given the input state and optimality. %However, the conventional policy network only performs direct amortization. 
While these schemes have improved the efficiency of VI by encoder networks \cite{kingma2014stochastic,rezende2014stochastic,mnih2014neural}, the output distribution of learned policy can be sub-optimal and far away from the target posterior, due to the insufficient expressivity of the policy network \cite{cremer2018inference,kim2018semi}. This suboptimality is typically defined as the amortization gap \cite{cremer2018inference}, resulting into a gap in the RL objective.

%In order to improve the RL performance, recent work has sought to reduce the amortization gap by transforming policy distribution through mappings with additional trainable weights to achieve richer posterior approximations \cite{agarwal2020imitative,rezende2015variational}, such as Normalizing Flow (NF) policy \cite{tang2018boosting,ward2019improving,mazoure2020leveraging}. 

%Although it demonstrates success in various RL domains, the NF policy does not explicitly use information about the target posterior of policy distribution. Hence, it is unclear whether the improved performance is resulted from better variational inference or simply the overparametrization of the sequence of transformations.

%In this work, inspired by the Hamiltonian Monte Carlo (HMC) technique and its use in VI \cite{caterini2018hamiltonian,zhang2018advances}, we propose to evolve actions from the base policy network according to HMC, which is to better approximate the posterior. %HMC is a popular MCMC technique, simulating a Markov chain in an extended space whose stationary distribution is the posterior, whereas VI chooses a family of tractable distributions and tries to find the member of that family with the lowest KL divergence to the posterior \cite{rezende2015variational,betancourt2017conceptual}.
The Hamiltonian Monte Carlo (HMC) has been used to improve VI in statistics \cite{caterini2018hamiltonian,zhang2018advances}. In this work, by leveraging the advantages of both VI and HMC \cite{salimans2015markov,wolf2016variational}, we propose to initialize Hamiltonian dynamics (HD) with samples from an optimized variational distribution, so that we can break the expressive limitation of the variational distribution and hence fill in the amortization gap. %Here we focus, in particular, on the use of time-inhomogeneous Hamiltonian dynamics, proposed originally in \cite{neal2011mcmc}.
Specifically, we propose to use HD to evolve the actions sampled from the policy network, so as to better approximate the target posterior and sample the actions with higher Q values, improving the efficiency of the exploration. % Rather than directly applying HD into RL, we propose a new leapfrog operator to simulate HD, which improves the expressivity of the policy distribution and  %We also use Hamiltonian dynamics with proposed leapfrog to improve the exploration efficiency, reaching a good balance of exploration and exploitation. 
We call this new policy integrated with HD as {\em Hamiltonian policy}. The proposed method offers several benefits.
%The conventional Gaussian policies in SAC \cite{haarnoja2018soft} are directionally uninformed, where actions are sampled in arbitrary directions with almost equal probabilities, wasting a lot of samples. And the expressivity of Gaussian policies is quite limited. However, 
First, the gradient information in Hamiltonian policy can make the exploration more directionally informed, avoiding sampling too many actions in opposite directions. Moreover, the randomness of momentum vectors in HD can help sampled actions to jump over the local optima and make the agent to explore more unknown parts of the state space. Further, the proposed leapfrog operator in Hamiltonian policy, which generalizes HMC via gated neural networks, can also increase the expressivity of the base policy network and adapt to the changing target distribution defined by Q function. Finally, in safe RL tasks, we find that the Hamiltonian policy can not only improve the achieved return by boosting the exploration, but also reduce the safety constraint violations by discarding potentially unsafe actions according to Lyapunov constraints \cite{chow2018lyapunov,chow2019lyapunov}.

Using empirical experiments, we evaluated the proposed method across a variety of benchmark continuous control tasks such as OpenAI Gym using the MuJoCo simulator \cite{todorov2012mujoco} and the realistic PyBullet Roboschool tasks \cite{coumans2016pybullet}. We show that the proposed method improves upon representative previous methods such as SAC \cite{haarnoja2018latent} and SAC with normalizing flow policy \cite{mazoure2020leveraging}, achieving both a better convergence rate and expected return. Additionally, we also empirically verify the advantage of our method in safe RL problems. 

In experiments, we conduct ablation study of the proposed leapfrog and sensitivity analysis on hyper-parameters. Additionally, we also compare the proposed method with iterative amortization policy optimization \cite{marino2020iterative}. And the action distribution of Hamiltonian policy is also visualized to show the improvement of expressivity.

\section{Preliminary}
In this section, we are going to introduce reinforcement learning (RL) as an Markov Decision Process (MDP). Then the constrained MDP and the solution based on Lagrangian method are introduced. We also formulate the RL problem in the framework of variational inference. Finally we briefly review the Soft Actor-Critic (SAC) \cite{haarnoja2018soft} and Hamiltonian Monte Carlo (HMC) \cite{neal2011mcmc} as building blocks of the proposed method.

\subsection{Markov Decision Process}
We consider Markov decision processes (MDP) as $(\mathcal{S}, \mathcal{A}, p_{\text{env}}, r)$, where $s_t\in\mathcal{S}$ and $a_t\in\mathcal{A}$ are the state and action at time step $t$, with the corresponding reward $r_t=r(s_t, a_t)$. The state transition of the environment is governed by $s_{t+1}\sim p_{\text{env}}(s_{t+1}|s_t, a_t)$, and the action is sampled from the policy distribution, given by the policy network $\pi_{\theta}(a_t|s_t)$ with parameters $\theta$. The discounted sum of rewards is denoted as $\mathcal{R}(\tau)=\sum_t\gamma^tr_t$, where $\gamma\in(0,1]$ is the discounted factor, and $\tau=(s_1,a_1,\ldots)$ is a trajectory. Thus, the distribution over the trajectory is
\begin{equation}
    p(\tau)=\rho(s_1)\prod_{t=1}^Tp_{\text{env}}(s_{t+1}|s_t,a_t)\pi_{\theta}(a_t|s_t) \label{traj_prob}
\end{equation}
where the initial state is drawn from the distribution $\rho(s_1)$. The objective of RL is to maximize the expected discounted return $\mathbb{E}_{p(\tau)}[\mathcal{R}(\tau)]$. 
At a given time step $t$, one can optimize this objective by estimating the accumulated future returns in the summation using an action-value network \cite{mnih2014neural,haarnoja2018soft}, denoted as $Q_{\pi}(s,a)$ in terms of a policy $\pi$. 

\subsection{Constrained MDP and Lagrangian Method}
Safety is an important issue in RL problems. We use constrained MDP (CMDP) to model RL problems in which there are constraints on the cumulative cost. The CMDP extends MDP by introducing a safety cost function and the associated constrained threshold, which is defined as $(\mathcal{S}, \mathcal{A}, p_{\text{env}}, r, c, d_0)$ where $c(s)\in[0,C_{\text{max}}]$ is a state-dependent cost function and $d_0\in\mathbb{R}_{>0}$ is an upper-bound on the expected cumulative safety cost in one episode. In addition to $Q_{\pi}$, we use another action-value network to approximate the accumulated future safety costs $\mathcal{C}(\gamma):=\mathbb{E}[\sum_t\gamma^t c_t]$, denoted as $Q_{C,\pi}$.

The Lagrangian method is a straightforward method to solve CMDP, by transforming it to a penalty form, i.e., $\max_{\theta}\min_{\lambda}\mathbb{E}[\sum_t\gamma^t(r(s_t,a_t)-\lambda c(s_t)|\pi_{\theta},s_0]$. The parameters $\theta$ and $\lambda$ are jointly optimized to a saddle-point. The policy parameters $\theta$ are optimized by a policy gradient algorithm, while the multiplier $\lambda$ is updated iteratively as $\lambda\longleftarrow[\lambda+\eta(J_C^{\pi}-d_0)]_+$, where $J_C^{\pi}$ is the discounted sum (or average sum) of safety costs in previous episodes.

\subsection{Reinforcement Learning via Variational Inference}
\label{sec:vi}
Recently a surge of works have formulated reinforcement learning and control as probabilistic inference \cite{dayan1997using,toussaint2006probabilistic,todorov2008general,botvinick2012planning,levine2018reinforcement}. In these works, the agent-environment interaction process is formulated as a probabilistic graphical model, then reward maximization is converted into maximum marginal likelihood estimation, where the policy resulting the maximal reward is learned via probabilistic inference. This conversion is accomplished by introducing one or more binary optimality variables $\mathcal{O}$. 
%, whose probability conditioned on the trajectory is proportional to the exponential of the return of that trajectory. 
%By referring variables $\mathcal{O}$ as optimality \cite{levine2018reinforcement}, the target is to learn the policy $\pi_{\theta}$ which can produce actions maximizing the likelihood of optimality. However evaluating this likelihood, i.e., $p(\mathcal{O}=1)=\int p(\mathcal{O}=1|\tau)p(\tau)d\tau$, needs the averaging over all the trajectories, which is computationally intractable especially in high-dimensions. Hence 
Since calculating the likelihood of optimality $\mathcal{O}$ requires intractable integral over all the possible trajectories, variational inference (VI) is adopted to lower bound the objective, where a variational distribution $q(\tau|\mathcal{O})$ is learned to approximate the posterior of trajectory given the optimality, yielding the evidence lower bound (ELBO) \cite{levine2018reinforcement}. The ELBO of the likelihood of optimality $\mathcal{O}$ can be written as below,
\begin{eqnarray}
    \lefteqn{\log p(\mathcal{O}=1)} \nonumber \\
    &&\ge\int q(\tau|\mathcal{O})\bigg[\log p(\mathcal{O}=1|\tau)+\log\frac{p(\tau)}{q(\tau|\mathcal{O})}\bigg]d\tau \nonumber \\
    &&=\mathbb{E}_{q}[\mathcal{R}(\tau)/\alpha]-D_{\text{KL}}(q(\tau|\mathcal{O})\|p(\tau))) \label{elbo}
\end{eqnarray}
where $D_{\text{KL}}(\cdot\|\cdot)$ denotes the KL divergence. Only model-free RL is considered here. We can simplify the EBLO in \eqref{elbo} by cancelling the probability of environmental dynamics. Then we can get the objective of policy optimization as below \cite{levine2018reinforcement},
\begin{equation}
    \mathcal{J}(q, \theta)=\mathbb{E}_{(s_t,r_t)\in\tau,\\ a_t\sim q}\bigg[\sum_{t=1}^T\gamma^tr_t-\alpha\log\frac{q(a_t|s_t,\mathcal{O})}{\pi_{\theta}(a_t|s_t)}\bigg] \label{obj}
\end{equation}
Specifically, at time step $t$, this objective can be written as
\begin{equation}
    \mathcal{J}(q,\theta)=\mathbb{E}_{q}[Q_{q}(s_t,a_t)]-\alpha D_{\text{KL}}(q(a_t|s_t,\mathcal{O})\|\pi_{\theta}(a_t|s_t)) \label{objective}
\end{equation}
Hence, with $\pi_{\theta}$ as action prior, policy optimization in the framework of VI \cite{haarnoja2018soft,levine2018reinforcement} is to find optimal $q$ maximizing the objective $\mathcal{J}(q,\theta)$ in \eqref{objective}. 

\subsection{Soft Actor-Critic}
\label{sec:sac}
Soft Actor-Critic (SAC) \cite{haarnoja2018soft} is a state-of-art off-policy RL algorithm widely used in many applications, especially in robotic problems with continuous actions and states. 
%It updates the policy using gradient descent, minimizing the KL divergence between the policy and the Boltzmann distribution using the Q-function as its negative energy function. 
SAC can also be formulated from the perspective of variational inference. When using uniform distribution $\mathcal{U}=(-1, 1)$ as the action prior $\pi_{\theta}$ in \eqref{objective}, the objective of SAC can be formulated as the state-action value function regularized with a maximum entropy,
\begin{equation}
\mathcal{L}(q) = \mathbb{E}_{s_t\sim\rho_{q}}\big[\mathbb{E}_{a_t\sim q}Q_q(s_t,a_t)-\alpha\log q(a_t|s_t)\big]. \label{sac_obj}
\end{equation}
where $q$ is the variational distribution of action. Here $\rho_q$ is the state distribution induced by policy $q$, and $\alpha$ is the temperature parameter which is introduced to improve the exploration. In this work, we are going to build the proposed method upon SAC. The optimal solution of \eqref{sac_obj} is $\bar{p}_{\alpha}(a|s)\propto\exp(Q(s,a)/\alpha)$ which is also the target policy distribution.

\subsection{Hamiltonian Monte Carlo}
\label{sec:hmc}
Hamiltonian Monte Carlo (HMC) is a popular Markov chain Monte Carlo (MCMC) method for generating sequence of samples, which converge to being distributed according to the target distribution \cite{neal2011mcmc}. Inspired by physics, the key idea of HMC is to propose new points by simulating the dynamics of a frictionless particle on a potential energy landscape $U(x)$ induced by a desired target distribution $p(x)$, where $p(x)\propto\exp(-U(x))$. This simulation is done in the formulation of {\em Hamiltonian dynamics} (HD). 
Specifically, HD is a reformulation of physical dynamics whose states can be described by a pair $(x, v)$ of $d$-dimensional vectors, where $x$ is the position vector and $v$ is the momentum vector. The dynamics of the system over time, i.e., the HD, is described by the Hamiltonian equations:
\begin{equation}
\frac{dx}{dt}=\frac{dH}{dx},\hspace{20pt}\frac{dv}{dt}=-\frac{dH}{dv} \label{hameq}
\end{equation}
where $H(x, v)$ is the Hamiltonian of the system, defined as the total energy of the system. In the physical context of HMC, the motion of the frictionless particle is governed by the potential energy $U(x)$ and kinetic energy $K(v)$. Since the Hamiltonian is the total energy here, we have $H(x,v)=U(x)+K(v)$, which is independent of time step due to the conservation of energy. The kinetic energy can be described as $K(v)=\beta v^Tv/2$ where $\beta$ is the mass of the particle, and the momentum vector is distributed as $p(v)\propto\exp(-\beta v^Tv/2)$ \cite{wolf2016variational}. 

The analytic solutions of HD \eqref{hameq} can determine three important properties of HMC algorithm, i.e., {\em reversibility, volume preservation} and {\em Hamiltonian conservation}. The reversibility means that the mapping $T_s$ from the state $(x_t,v_t)$ at time $t$ to some future state at time $t+s (s > 0)$ is one-to-one and reversible. The volume preservation implies that the transformation based on HD conserves the volume in state space, i.e., applying $T_s$ to some region results in another region with the same volume. Finally, the Hamiltonian $H(x,v)$ stays constant with time, i.e., $dH/dt=0$, which is called Hamiltonian conservation.

The HD described in \eqref{hameq} is typically simulated by the {\em leapfrog operator} \cite{leimkuhler2004simulating,neal2011mcmc}, of which the single time step can be described as
\begin{equation}
v^{\frac{1}{2}}=v-\frac{\epsilon}{2}\partial_x U(x);\hspace{5pt} x'=x+\epsilon v^{\frac{1}{2}};\hspace{5pt} v'=v^{\frac{1}{2}}-\frac{\epsilon}{2}\partial_{x'} U(x'); \label{org_leapfrog}
\end{equation}
which transforms $(x,v)$ to $(x',v')$. We can see that transformations in \eqref{org_leapfrog} are all volume-preserving shear transformations, where in every step only one of variables ($x$ or $v$) changes, by an amount determined by the other one. Hence the Jacobian determinant of \eqref{org_leapfrog} is simply 1 and the density of transformed distribution $p(x',v')$ is tractable to compute.
%In this work we focus on the time-inhomogeneous HD \cite{neal2011mcmc,caterini2018hamiltonian}. This method uses reverse kernels which are optimal for reducing variance of the likelihood estimators and allows for simple calculation of the approximate posteriors. 

%\subsection{Optimal Control}
%Here we regard the policy optimization iterations as system dynamics subject to an external influence, or control. A standard formulation for discrete-time systems with feedback control is:
%\begin{eqnarray}
%\bm{x}_{k+1}&=&F(\bm{x}_k, \bm{u}_k) \nonumber \\
%\bm{y}_k&=&Z(\bm{x}_k) \nonumber \\
%\bm{u}_k&=&h(\bm{y}_0,\ldots,\bm{y}_k) \nonumber
%\end{eqnarray}
%with state vector $\bm{x}$, dynamics function $F$, measurement outputs $\bm{y}$, applied control $\bm{u}$, and the subscript denoting the time step. The feedback rule $h$ has access to past and present measurements. The goal in optimal control is to design a control rule, $h$, that results in a sequence $\bm{y}_{0:T}=\{\bm{y}_0,\ldots,\bm{y}_T\}$, scoring highly according to some metric $C$.

%It is always easier to analyze and control systems with simpler dependence on the input, even though the dependence on the state is complex. Control-affine systems are a broad class of dynamical systems which are especially amenable to analysis \cite{isidori2013nonlinear}. Generally the dynamics there take the form
%\begin{equation}
%    F(\bm{x}_k, \bm{u}_k) = f(\bm{x}_k) + g(\bm{x}_k)\bm{u}_k %\nonumber
%\end{equation}
%where $f$ and $g$ can be non-linear and unknown a prior.
\section{Related Work}
There have been a lot of previous works on improve policy optimization in recent years. To optimize the Q-value estimator with an iterative derivative-free optimizer, Qt-opt \cite{kalashnikov2018qt} uses the cross-entropy method (CEM) \cite{rubinstein2013cross} to train robots to grasp things. To improve the model-predictive control, CEM and related methods are also used in model-based RL \cite{nagabandi2018neural,chua2018deep,piche2018probabilistic,schrittwieser2020mastering}. %Entropy regularization is another active field in policy optimization \cite{mnih2016asynchronous,nachum2017bridging,haarnoja2018soft,haarnoja2017reinforcement}. Recent work introduces entropy regularization into approximate policy iteration approaches \cite{haarnoja2018soft,yang2019regularized}. The authors in \cite{ahmed2019understanding} show the impact of entropy regularization and empirically the effectiveness of the entropy regularization on smoothing the optimization landscape.

However, there are less recent works on gradient-based policy optimization \cite{henaff2017model,srinivas2018universal,bharadhwaj2020model,marino2020iterative}. They are specifically designed for model-based RL \cite{henaff2017model,srinivas2018universal,bharadhwaj2020model}. Normalizing flow \cite{haarnoja2018latent,tang2018boosting,mazoure2020leveraging} is another method to improve the policy optimization, by increasing the expressivity of the policy network. But none of them include gradient information, so that exploration is not sufficient in some environments. Another significant challenge with this approach is the Jacobian determinant in the objective, which is generally expensive to compute. Previous methods make the Jacobian determinant easy-to-evaluate at the sacrifice of the expressivity of the transformation, where the determinant only depends on the diagonal \cite{kingma2016improved,tang2018boosting,mazoure2020leveraging}, limiting the exploration in the RL process.  

Another approach is to apply iterative amortization in policy optimization, which uses gradients of Q function to iteratively update the parameters of the policy distribution \cite{haarnoja2017reinforcement,marino2020iterative}. However, especially when the estimation bias of Q functions is significant \cite{ciosek2019better}, directly using gradients to improve policy distributions without additional randomness may make the policy search stuck at local optima which limits the exploration, and the policy distribution therein is still Gaussian. That is why methods in \cite{marino2020iterative} cannot outperform SAC in many MuJoCo environments. Finally, one more related work is Optimistic Actor Critic (OAC) \cite{ciosek2019better} using gradient of value function to update actions in exploration. However, their updated policy distribution is still Gaussian which has limited expressivity and one-step update with gradient is not enough to have significant performance improvement.

%% file: part2.tex
\section{Methodology}
%Based on the empirical success of HMC \cite{neal2011mcmc,hoffman2014no}, many algorithms exploiting Hamiltonian dynamics (HD) have been developed to obtain unbiased estimates of the ELBO and its gradients \cite{salimans2015markov,wolf2016variational,caterini2018hamiltonian}. %For the policy optimization in the framework of VI, the key step is to estimate the gradient of ELBO \eqref{obj} with low variance. 
%In this work, we propose to use HD to evolve actions produced by the base policy network, so as to make the policy distribution better approximate the posterior of action given the state and optimality, i.e., the maximizer of the ELBO \eqref{obj}. 
%Additionally, we can easily use the reparameterization trick to calculate unbiased gradients of the ELBO with respect to the parameters of interest.

In this section, we first formulate the proposed policy optimization method in the framework of variational inference (VI), and then propose a new leapfrog operator to quickly adapt to the changes of the target distribution during the learning. Finally implementation considerations and its application into safe RL are introduced.
%In this work, we propose to build Hamiltonian dynamics on top of the policy network, so as to reduce the amortization gap, improve the expressivity of the policy distribution, and make the exploration more directionally informed.
%Reinforcement learning can be formulated as a probabilistic graphic model (PGM) and solved by probabilistic or variational inference methods, where the policy network produces actions or policy distributions to approximate the posterior conditioned on the optimality $\mathcal{O}$ \cite{levine2018reinforcement}. Hence, the policy network is essentially the variational distribution, playing the same role as the encoder in variational autoencoder (VAE) \cite{kingma2014stochastic}. According to the amortization gap \cite{cremer2018inference} widely observed in VAE, . In this work, in order to close this gap, we propose to build HMC iterations on top of the policy network. Specifically, Hamiltonian dynamics are initialized by the output of policy network, and the optimization of parameters in HMC is incorporated into the optimization of policy network parameters. Moreover, HMC can also make the exploration more directionally informed, improving the data efficiency significantly.

\subsection{Hamiltonian Policy Optimization}
We call the method of using Hamiltonian policy into RL as Hamiltonian policy optimization (HPO). A feature of HPO is that a momentum vector $\rho$ is introduced to pair with the action $a$ in dimension $d_a$, extending the Markov chain to work in a state space $(a,\rho)\in\mathbb{R}^{d_a}\times\mathbb{R}^{d_a}$. %The randomness of momentum vectors here can help actions jump out of local optima, playing important roles in policy optimization and exploration. 
Specifically, the momentum vector $\rho$ has Gaussian prior $\mathcal{N}(\rho|0,\beta_0^{-1} I)$, and the action $a$ follows the uniform prior $\mathcal{U}(-1, 1)^{d_a}$, where $\beta_0$ is a hyper-parameter which determines variance of $\rho$.
%output distribution of the last updated policy $\pi_{\theta_t}(a|s)$ for iteration $t$ of policy optimization. 
According to the discussion in Section \ref{sec:vi} and \ref{sec:sac}, the target distribution in HPO, i.e., target posterior of action and momentum vector, can be written as
\begin{equation}
    \bar{p}_{\alpha}(a,\rho|s)\propto\exp\big(Q_{\pi_{\theta}}(s,a)/\alpha\big)\mathcal{N}(\rho|0,\beta_0^{-1} I) \label{target}
\end{equation}
Therefore, following Section \ref{sec:hmc}, the target potential function and momentum kinetic energy can be written as $U_{\theta}(s, a):=-Q_{\pi_{\theta}}(s,a)/\alpha$ and $K(\rho):=\frac{\beta_0}{2}\rho^T\rho$. 

The core innovation of HPO is to evolve $a$ and $\rho$ via Hamiltonian dynamics (HD) in \eqref{hameq}, where HD is approximated by steps of deterministic transitions (leapfrog in \eqref{org_leapfrog}), so that the evolved actions would likely better approximate the target posterior. %and hence decrease the amortization gap \cite{cremer2018inference}. %Denote one step of Hamiltonian dynamic as $\Phi^k_{\theta, h}(\cdot,\cdot)$, where $\theta$ denotes the parameters of policy $\pi_{\theta}$ in \eqref{target}, and 
%Essentially we evolve the action and momentum vector via deterministic transitions $q^k_{\theta, h}(a_k,\rho_k|a_{k-1},\rho_{k-1})=\delta_{\Phi^k_{\theta, h}(a_{k-1},\rho_{k-1})}(a_k,\rho_k)$, where $\theta$ denotes the parameters of policy $\pi_{\theta}$ in \eqref{target} and 
%$h$ represents trainable parameters in simulating HD, so that we can have 
%\begin{equation}
%(a_K,\rho_K)=f^K_{\text{HMC}}(a_0,\rho_0;\theta, h):=\big(\Phi^K_{\theta,h}\circ\cdots\circ\Phi^1_{\theta,h}\big)(a_0,\rho_0)  \label{hmcoperation}  
%\end{equation}
%where $(a_0,\rho_0)\sim\pi_{\theta}(\cdot|s)\mathcal{N}(0,\beta_0^{-1}I)$, and $\{\Phi^k_{\theta,h}\}_{k=1}^K$ define diffeomorphisms corresponding to a time-discretized and inhomogeneous HD \cite{caterini2018hamiltonian}.
%Different from normalizing flow policies in previous works \cite{tang2018boosting,ward2019improving,mazoure2020leveraging}, the gradient information of Q-function is incorporated, and the log density of output actions is easy to compute due to the tractable Jabcobian determinant, facilitating the policy entropy regularization in policy optimization. 

\begin{figure*}
    \centering
    \subfigure[Hamiltonian Policy]{
        \centering
        \fontsize{8pt}{10pt}\selectfont
        \def\svgwidth{0.8\columnwidth}
        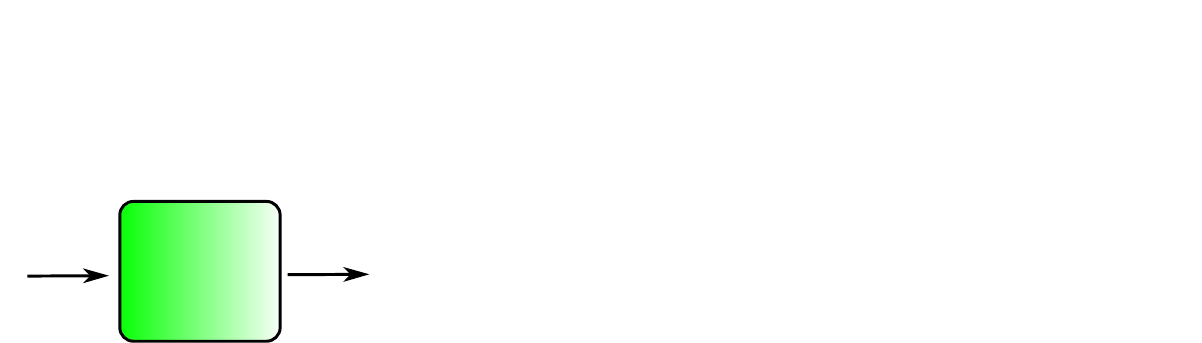
        \label{fig:pol}
    }
    \subfigure[Proposed Leapfrog]{
    \fontsize{7pt}{10pt}\selectfont
    \def\svgwidth{0.89\columnwidth}
    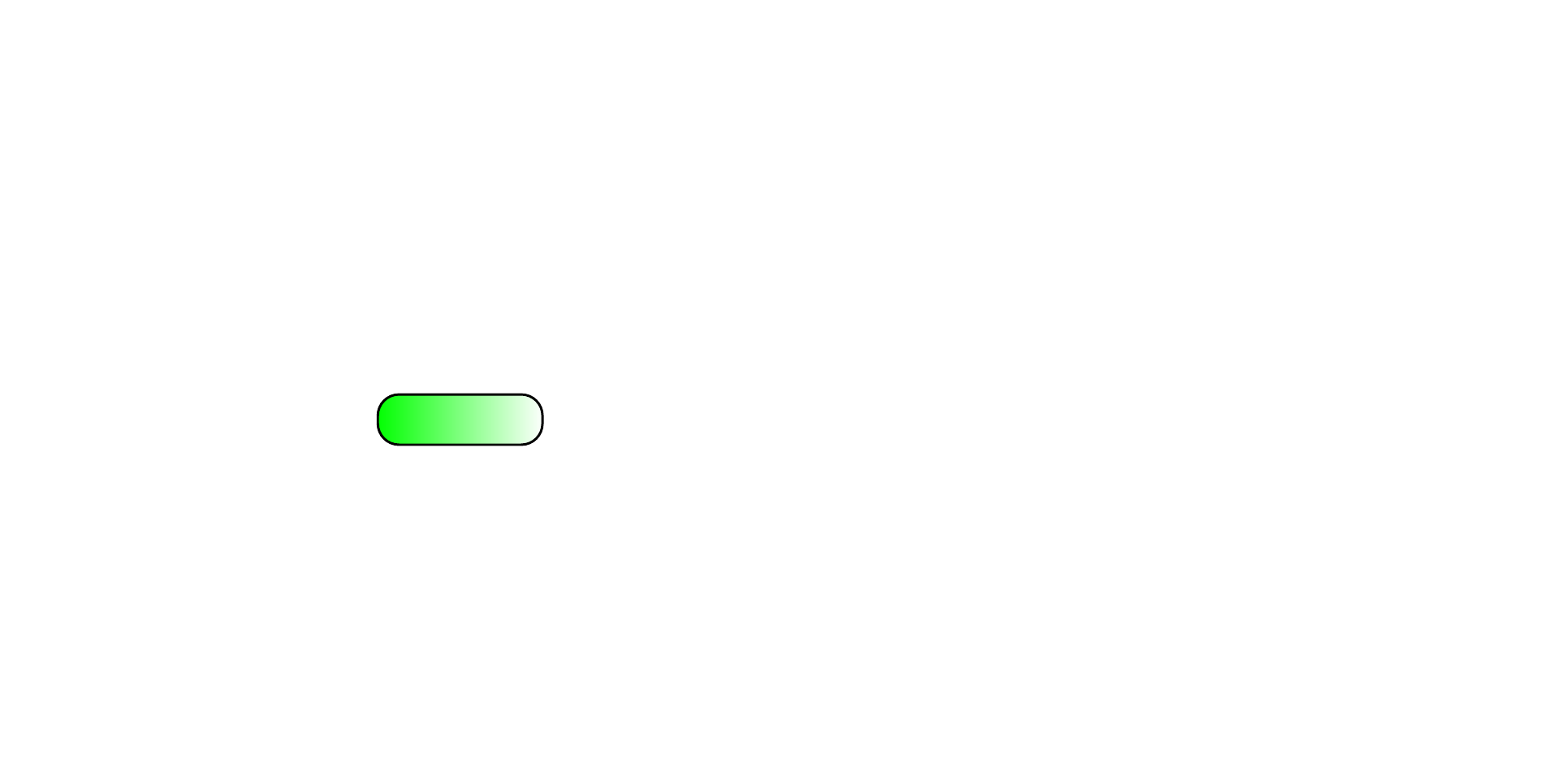
    \label{fig:hmc}
    }
    \caption{Diagram of the Hamiltonian policy and proposed leapfrog operator. The HMC box represents one step leapfrog operator.}
    \label{fig:proposed}
\end{figure*}

%Every transition $\Phi^k_{\theta,h}$ here consists of two steps: a leapfrog step, which discretizes the HD, and a tempering step, which adds inhomogeneity to the dynamics and allows us to explore isolated modes of the target distribution \cite{caterini2018hamiltonian,neal2011mcmc}. The leapfrog transforming $(a,\rho)$ to $(a',\rho')$ can be described as the following transformations:
In Hamiltonian policy, given input state $s$, the initial action $a_0$ and momentum vector $\rho_0$ are sampled as $(a_0,\rho_0)\sim\pi_{\theta}(\cdot|s)\mathcal{N}(0,\beta_0^{-1}I)$ where $\pi_{\theta}$ is the base policy network. Then, by using leapfrog operator in \eqref{org_leapfrog} to simulate HD, since $U_{\theta}(s, a):=-Q_{\pi_{\theta}}(s,a)/\alpha$, action and momentum are evolved iteratively by the leapfrog (HMC step) as below,
%Every transition $\Phi^k_{\theta,h}$ here consists of a leapfrog step, which discretizes the Hamiltonian dynamics. The leapfrog transforming $(a,\rho)$ to $(a',\rho')$ can be described as the following transformations:
\begin{eqnarray}
\rho_{k+1/2}&=&\rho_k+\frac{\epsilon}{2}\odot\nabla Q_{\pi_{\theta}}(s, a_k)/\alpha \nonumber \\
a_{k+1}&=&a_k + \epsilon\odot\rho_{k+1/2} \nonumber \\
\rho_{k+1}&=&\rho_{k+1/2}+\frac{\epsilon}{2}\odot\nabla Q_{\pi_{\theta}}(s, a_{k+1})/\alpha\label{leapfrog}
\end{eqnarray}
where $\nabla$ is the differentiation taken with respect to $a$, $\epsilon\in\mathbb{R}$ is the learning step size, and $k=0,\ldots,K-1$. Then by evolving initial action $a_0$ for $K$ leapfrog steps, the action $a_K$ is applied to the environment finally. The working process is shown in Figure \ref{fig:pol}

Denote the $k$-th leapfrog step described above as $(a_k, \rho_k):=\Phi^k_{\theta,h}(a_{k-1},\rho_{k-1})$. We can see that each leapfrog step still has unit Jacobian. %Thus the leapfrog step can be represented as $(a_k,\rho_k)=\Phi^k_{\theta,h}(a_{k-1},\rho_{k-1})$. Denote the distribution of action and momentum at $k$-th step of transitions as $q^k_{\theta,h}(a_k,\rho_k)$. And since the initial action and momentum are sampled from variational distributions, we have  $q^0_{\theta,h}(a_0,\rho_0):=\pi_{\theta}(a|s)\mathcal{N}(\rho|0,I)$. 
%Thus the process of transforming $(a_0,\rho_0)$ to $(a_K,\rho_K)$ can be written as 
%For the tempering step, the momentum output $\rho'$ of each leapfrog step is multiplied by a scalar $\alpha_k\in(0, 1)$ for $k=1,\ldots,K$. Since the target distribution defined by Q-function is varying during the learning process and Q-function may have estimation bias, we adopt fixed tempering scheme proposed in \cite{caterini2018hamiltonian}, which can keep the learning process numerically stable. Specifically $\alpha_k=\sqrt{\beta_{k-1}/\beta_k}$ and $\beta_k$ is a deterministic function of $\beta_0$ shown in \eqref{beta_k}. Then we can easily have that the Jacobian of every transition $\Phi^{k}_{\theta,h}$ is given by $|\det\nabla\Phi^{k}_{\theta,h}(a_k,\rho_k)|=\alpha_k^{d_a}=(\beta_{k-1}/\beta_k)^{d_a/2}$.
Therefore, based on the change of variable formula in probability distribution, the joint distribution of action and momentum variables after $K$ steps of leapfrog can be expressed as
\begin{eqnarray}
    q^K_{\theta,h}(a_K,\rho_K)&=&q^0_{\theta,h}(a_0,\rho_0)\prod_{k=0}^{K-1}\big|\det\nabla\Phi^{k+1}_{\theta,h}(a_k,\rho_k)\big|^{-1} \nonumber \\
    %&=&q^0_{\theta,h}(a_0,\rho_0)\prod_{k=1}^{K}\bigg(\frac{\beta_{k-1}}{\beta_k}\bigg)^{-d_a/2} \nonumber \\
    &=&\pi_{\theta}(a_0|s)\mathcal{N}(\rho_0|0,\beta_0^{-1}I)  \label{hmcprob} %\beta_0^{-d_a/2} 
\end{eqnarray}
where $(a_K,\rho_K)$ are action and momentum evolved by $K$ HMC steps. Hence the density of output action and momentum vector becomes tractable to compute, facilitating the policy entropy regularization in SAC-style algorithms. 

In the framework of VI, the policy optimization objective of HPO is the ELBO \eqref{elbo}. Since ELBO can be written as the difference between the log of target distribution and log of variational distribution \cite{kingma2014stochastic}, we can write the ELBO for HPO as below,
\begin{equation}
\mathcal{L}_{\text{ELBO}}(\theta,h;s) = \mathbb{E}_{(a_0,\rho_0)}[\log\bar{p}_{\alpha}(a_K,\rho_K|s)-\log q^K_{\theta,h}(a_K,\rho_K)] \label{hmcelbo}
\end{equation} 
The policy network parameters are denoted as $\theta$ and parameters in HMC are denoted as $h$.

Finally, combining \eqref{target}, \eqref{hmcprob} and \eqref{hmcelbo} together and ignoring terms not related with $\theta$ and $h$, the objective of HPO, i.e., the expectation of EBLO over all the visited states, can be written as
\begin{equation}
    \mathcal{J}(\theta,h)=\mathbb{E}_{s\sim \rho_{\pi_{\theta}}}\bigg[Q_{\pi_{\theta}}(s,a_K) - \alpha \log{\pi_{\theta}(a_0|s)} - \frac{\alpha\beta_0}{2}\rho^T_K\rho_K \bigg] \label{hmcobj}
\end{equation}
where $\rho_{\pi_{\theta}}$ is the state distribution induced by the policy $\pi_{\theta}$.
%$(a_0,\rho_0)\sim\pi_{\theta}(\cdot|s)\mathcal{N}(\cdot|0,\beta_0^{-1}I)$ and $(a_K,\rho_K)=f^K_{\text{HMC}}(a_0,\rho_0;\theta, h)$. 
Note that $\alpha$ is the temperature parameter tuned in the same way as SAC \cite{haarnoja2018soft}. And $Q_{\pi_{\theta}}$ is approximated by a target critic network which is not related with $\theta$ and is periodically updated in the learning process \cite{haarnoja2018latent}.

\subsection{Proposed Leapfrog Operator}
Since HMC with conventional leapfrog \eqref{leapfrog} converges and mixes slowly, some past works proposed to use neural networks to generalize HMC \cite{levine2018reinforcement,li2019neural}.
%the action value given by the Q network may have estimation bias, and directly incorporating gradient information like \eqref{leapfrog} may  mislead the learning to wrong regions in the policy space, leading to suboptimal policy estimate. 
%since the Q value is changing in the learning process, how to make the policy distribution quickly adapt to updated Q networks in a numerically stable way is the desiderata here. The key is to improve policy expressivity without scarifying numerical stability. T
However, since Q networks are changing in RL, the techniques proposed in \cite{levine2018reinforcement,li2019neural} cannot be used here. Based on our empirical study, the direction variable, binary mask and $\exp$ operation therein \cite{levine2018reinforcement,li2019neural} can make the policy optimization unstable, degrading the RL performance.

Instead, we propose to use a gating-based mechanism to generalize the conventional leapfrog operator \eqref{leapfrog} and design a new leapfrog operator, which integrates gradient information via both explicit and implicit approaches.
%, shown in Figure \ref{fig:hmc} in Appendix \ref{sec:diagram}. 
The explicit approach is to directly use the primitive gradient same as \eqref{leapfrog}, whereas the implicit approach is to use an MLP $T_h$ to transform the primitive gradient, state and action together. Then the gradient information from both explicit and implicit approaches are combined by a gate $\sigma_h$. The motivation behind is to improve the policy expressivity by MLP $T_h$ and control the numerical stability by gate $\sigma_h$, making the policy distribution quickly adapt to the changes of Q networks during the learning process. %So, optimizing networks $T_h$ and $\sigma_h$ together with the policy network can make the incorporated gradients adaptive to the policy optimization process, and hence improve the learning performance. 

%According to empirical experiments we find that the gradient normalized by its $l_2$ norm can make the model perform best. 
The inputs of $T_h$ and $\sigma_h$ include normalized gradients, action and state, where the state is optional and can be ignored in some environments.
%The output of $\sigma_h$ is a vector with each element bounded in $[0,1]$, controlling the trade-off between explicit and implicit gradients. 
Therefore, the proposed leapfrog operation, transforming from $(a_k, \rho_k)$ to $(a_{k+1},\rho_{k+1})$, can be described as 
\begin{eqnarray}
\rho_{k+1/2}&=&\rho_k-\frac{\epsilon}{2}\odot(\sigma_h(s,a_k,g)\odot g \nonumber \\
&&+ (1-\sigma_h(s,a_k,g))\odot T_h(s,a_k,g)) \nonumber \\
\rho_{k+1}&=&\rho_{k+1/2}-\frac{\epsilon}{2}\odot(\sigma_h(s,a_{k+1},g')\odot g' \nonumber \\
&&+ (1-\sigma_h(s,a_{k+1},g'))\odot T_h(s,a_{k+1},g'))\label{leapfrog1}
\end{eqnarray}
where $g:=\frac{-\nabla Q_{\theta}(s,a_k)}{\|\nabla Q_{\theta}(s,a_k)\|}, g':=\frac{-\nabla Q_{\theta}(s,a_{k+1})}{\|\nabla Q_{\theta}(s,a_{k+1})\|}$ and $a_{k+1}=a_k + \epsilon\odot\rho_{k+1/2}$. This process is shown in Figure \ref{fig:hmc}.  %and the output density of steps of the proposed leapfrog operator is tractable to compute. %, where the explicit and implicit gradients are combined by the gate $\sigma_h$. 
%Specifically, $T_h$ and $\sigma_h$ have different parameters contained in $h$, but they should have the same architecture based on empirical study. 
%We find that the architecture of $T_h$ and $\sigma_h$ should be simple, having one hidden layer with 32 or 64 units. The output activations of $T_h$ and $\sigma_h$ are sigmoid and tanh respectively. %The transformations in \eqref{leapfrog1} are visualized in Figure \ref{fig:hmc} in Appendix \ref{sec:diagram}. 
Obviously the proposed leapfrog operator \eqref{leapfrog1} still keeps the properties of reversibility and unit Jacobian, so that the distribution of $(a_K, \rho_K)$ in \eqref{hmcprob} is still tractable.
%During training, parameters $h$ are optimized with the parameters of the policy network $\theta$. 

\subsection{Implementation and Algorithms}
In implementation, we build the Hamiltonian dynamics (HD) simulated by leapfrog steps on top of the policy network. %The randomness of momentum vector $\rho$ can boost the exploration and improve the generalization of the policy trained in the policy optimization. The neural networks $T_h$ and $\sigma_h$ in the proposed leapfrog operator can increase the expressivity of policy distribution, verified in the experiment section. In exploration, the policy distribution is expanded towards regions with higher Q values, which makes the exploration more directed and hence improves the sample efficiency.
Specifically, we only use one hidden layer for the base policy network $\pi_{\theta}$, so the number of parameters of our model is much smaller than that of models in previous papers \cite{haarnoja2018latent,haarnoja2017reinforcement,marino2020iterative} which use two hidden layers in the policy network. 
%It shows that the performance improvement of our method is not from extra parameters introduced by $T_h$ and $\sigma_h$ in the leapfrog operators.
% Specifically, we find that the HD has different effects on performance in the exploration and policy optimization, which is analyzed in the experiment section. Using different momentum variances $\beta_0^{-1}$ for exploration and policy optimization, denoted as ${\beta_0^{\text{exp}}}^{-1}$ and ${\beta_0^{\text{tr}}}^{-1}$ respectively, can yield the best empirical performance. And in general we set ${\beta_0^{\text{exp}}}^{-1}>{\beta_0^{\text{tr}}}^{-1}$, since the exploration needs more randomness than policy optimization. 
%In order to achieve the best empirical performance, we make some specific changes in the implementation of RL algorithms. First, it is better to omit the last term of the objective \eqref{hmcobj}, i.e., $\frac{\alpha\beta_0}{2}\rho^T_K\rho_K$, in the practical implementation of policy optimization. Second, different from previous works on HMC \cite{salimans2015markov,wolf2016variational,levy2018generalizing}, the Metropolis-Hastings accept/reject step is omitted here. That is because the Q values are dramatically varying during the training of RL. If accept/reject step was applied here, the acceptance rate could be very low, which could limit the exploration and hence degrade the learning performance significantly. 
The proposed RL algorithm is built on top of SAC, where the Gaussian policy is replaced by Hamiltonian policy and the policy optimization objective in \eqref{hmcobj} is used. It is termed as SAC-HPO. The process of producing actions from Hamiltonian policy is shown in Figure \ref{fig:pol}. The details of the algorithm are presented in Algorithm \ref{alg:hpo} and Algorithm \ref{alg:hd}.

\begin{algorithm}[ht]
\caption{Hamiltonian Policy Optimization}
\label{alg:hpo}
\begin{algorithmic}[1]
\State {Denote $a_t, s_t$ as the action and state at time$t$; Denote the replay buffer as $\mathcal{B}$;}
\State Initialize $\theta, h$
\For{$t=1,2,\ldots$}
\State Sample $a_t\sim\pi_{\theta_t}(\cdot|s_t)$
\State Obtain $a^K_t, \rho^K_t=f^K_{\text{HMC}}(s_t, a_t; \theta_t, h_t)$
\State Apply $a^K_t$, and obtain next state $s_{t+1}$
\State Store the experience tuple $(s_t, a^K_t, s_{t+1})$ into $\mathcal{B}$
\State Sample a minibatch of transitions $\mathcal{D}_t$ from $\mathcal{B}$
\State Update the Q network by $\mathcal{D}_t$
%Obtain the evolved action and momentum $a^K, \rho^K = f^K_{\text{HMC}}(s, a; \theta_t, h_t), \forall (s, a)\in\mathcal{D}_t$\;
\State Update $\theta$ and $h$ by optimizing $\mathcal{J}(\theta, h)$ in \eqref{hmcobj}\ with minibatch $\mathcal{D}_t$
\EndFor
\end{algorithmic}
\end{algorithm}

\begin{algorithm}[ht]
\caption{$f^K_{\text{HMC}}(s, a; \theta, h), \beta_0, \epsilon$}
\label{alg:hd}
\begin{algorithmic}[1]
\State Sample $\rho^0\sim\mathcal{N}(0, I)$
\State Set $\rho_0\longleftarrow\rho_0\cdot\frac{1}{\sqrt{\beta_0}}$
\For{$k=1,\ldots,K$}
\State Obtain $\rho_{k+1/2}$ by the first equation in \eqref{leapfrog1}
\State Update $a_k=a_{k-1}+\epsilon\odot\rho_{k+1/2}$
\State Obtain $\rho_{k+1}$ by the second equation in \eqref{leapfrog1}
\EndFor
\State Return {$a_K, \rho_K$}
\end{algorithmic}
\end{algorithm}

\subsection{Safe Reinforcement Learning with Hamiltonian Policy}
\label{sec:saferl}
In addition to regular RL, we also find the proposed method can be used in safe RL to reduce the safety violations. It is based on Lyapunov constraint \cite{chow2018lyapunov,chow2019lyapunov}. Under this constraint, based on their assumptions, the updated policy is provably guaranteed to satisfy the safety constraint. It transforms the trajectory-wise safety constraint to a state-wise constraint \cite{chow2019lyapunov}. Specifically, the Lyapunov constraint is expressed as 
\begin{equation}
    Q_{C,\pi_B}(s,a) - Q_{C,\pi_B}(s,\pi_B(s)) < \tilde{\epsilon}  \label{lyapunov}
\end{equation}
where 
\begin{equation}
\tilde{\epsilon}=(1-\gamma)\cdot(d_0-Q_{C,\pi_B}(s_0,\pi_B(s_0))) 
\end{equation}
where $\pi_B$ is the reference policy which is the updated policy in last iteration, $Q_{C,\pi_B}$ is the accumulated safety costs in terms of policy $\pi_B$, $s_0$ is the initial state and $\gamma$ is the discounting factor of the MDP. 

However, in previous work \cite{chow2018lyapunov,chow2019lyapunov}, the sampled actions are made to satisfy the Lyapunov constraints based on the linear approximation of the cost critic $Q_{C,\pi_B}$, which is inaccurate in practice since the action may not have a linear relationship with the future costs. In this paper, we propose to use Hamiltonian policy to iteratively sample actions until the Lyapunov constraint is satisfied, so that potentially unsafe actions can be discarded in this process.

When interacting with the environment, for any sampled action, the agent first uses Lyapunov-constraint to predict its safety violation. If the Lyapunov-constraint is satisfied, it is applied into the environment. Otherwise, the sampled action will be updated by the leapfrog operator to get the next sampled actions, until the sampled actions satisfy the Lyapunov constraint. So, HMC here can not only boost the exploration by randomness and gradient information, but also improve safety in exploration by sampling actions iteratively until the safety (Lyapunov) constraint is satisfied. The application of Hamiltonian policy in safe RL is summarized in Algorithm \ref{alg:safe}. Note that compared with regular RL, a difference of Hamiltonian policy in safe RL is that the random noise is injected into momentum variables in every leapfrog step, rather than initial step only.

\begin{figure*}[ht]
\centering
\subfigure[HalfCheetah-v2]{
    \includegraphics[width=1.5in]{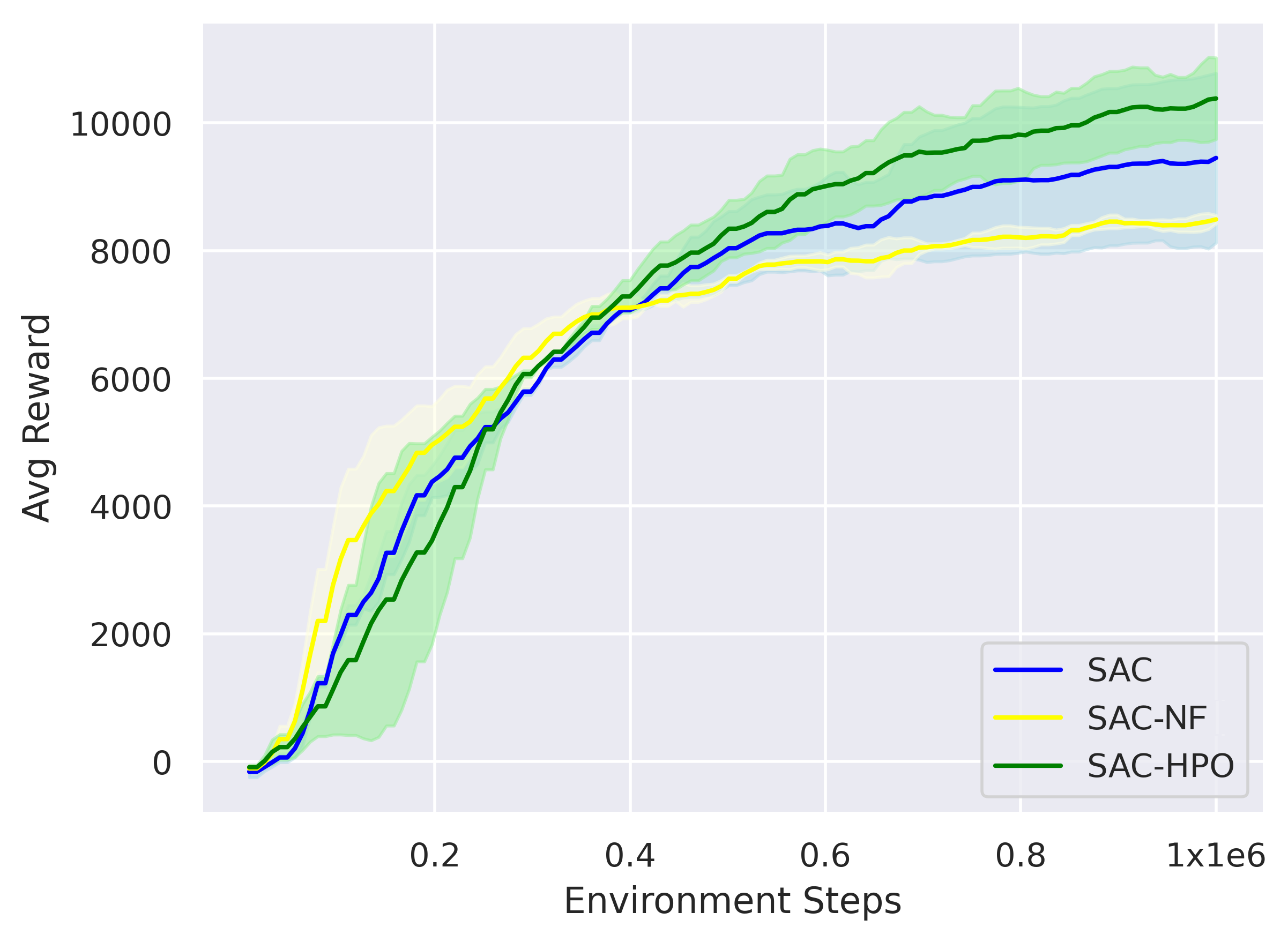}
    \label{fig:mujoco_1}
}
\subfigure[Walker2d-v2]{
    \includegraphics[width=1.5in]{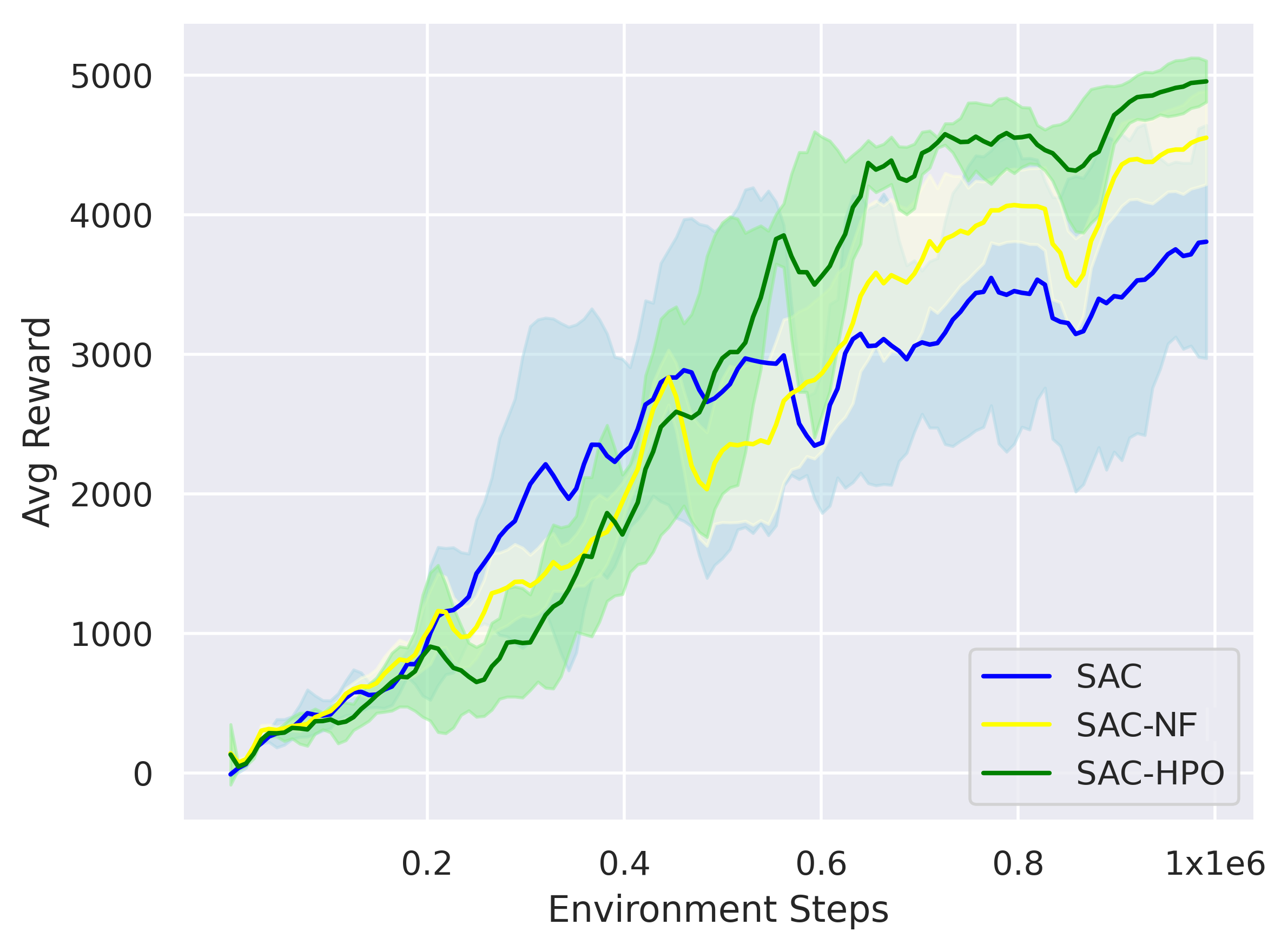}
    \label{fig:mujoco_2}
}
\subfigure[Hopper-v2]{
    \includegraphics[width=1.5in]{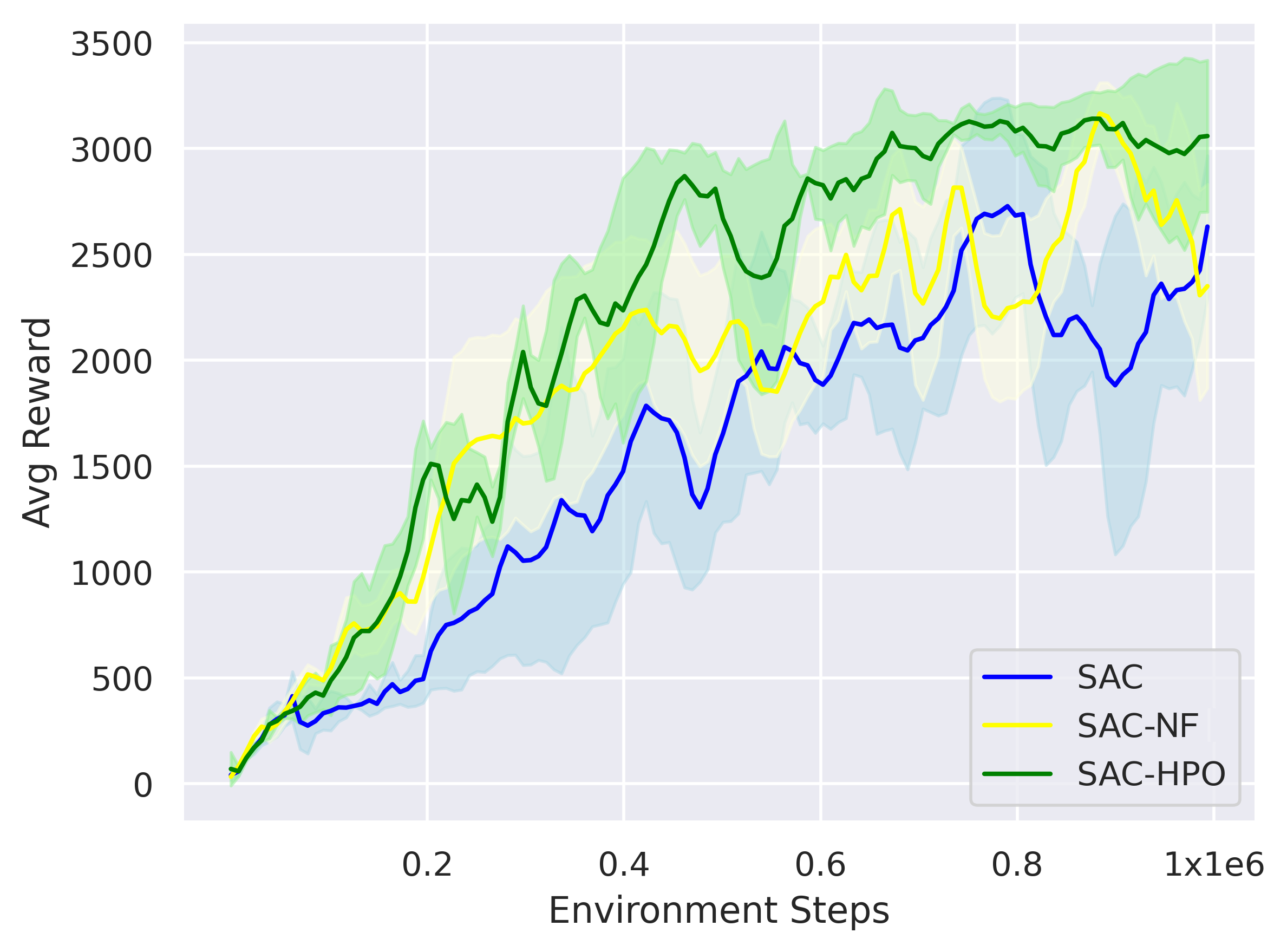}
    \label{fig:mujoco_4}
}
\subfigure[Ant-v2]{
    \includegraphics[width=1.5in]{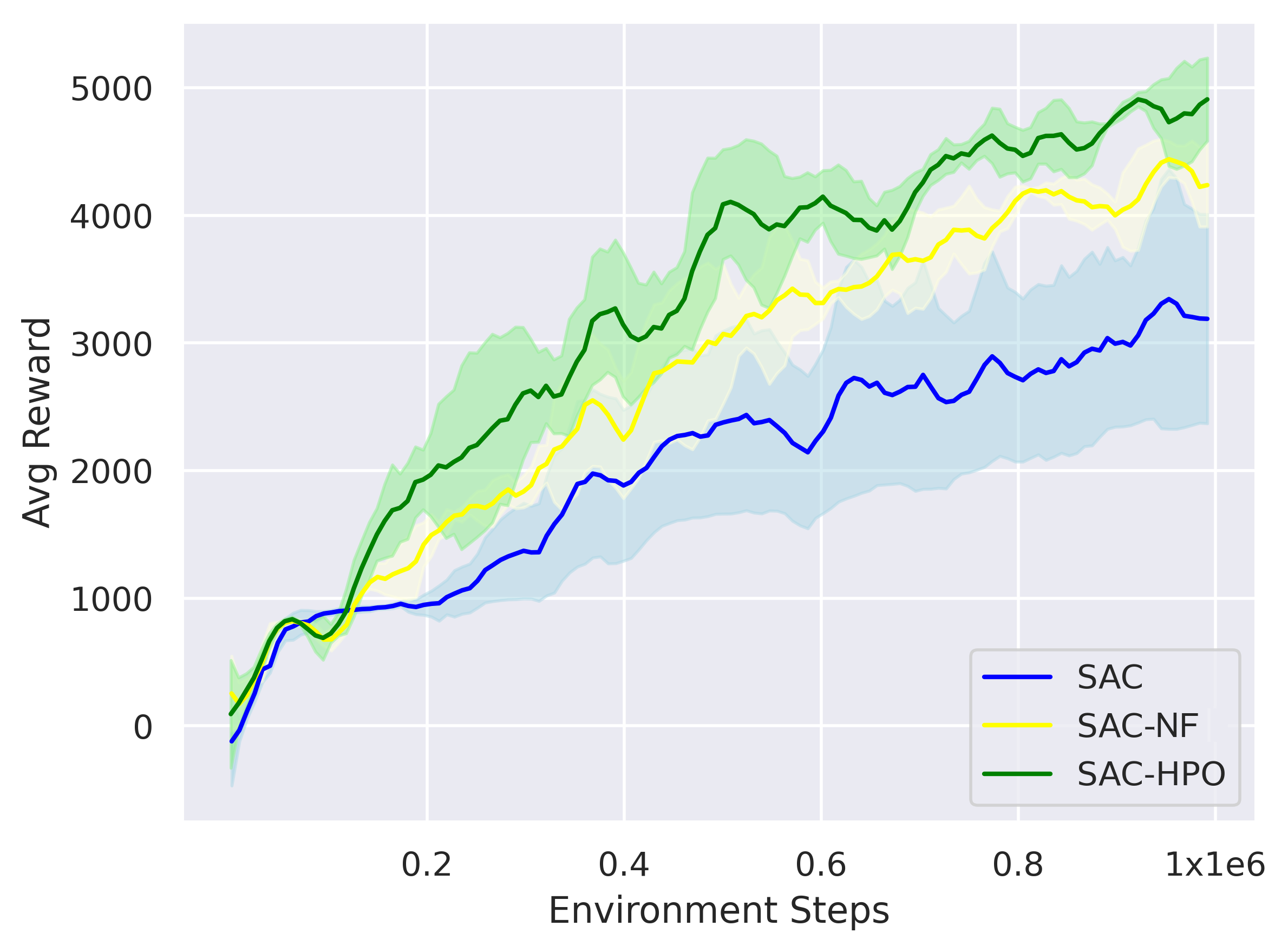}
    \label{fig:mujoco_3}
}

\subfigure[Humanoid-v2]{
    \includegraphics[width=1.5in]{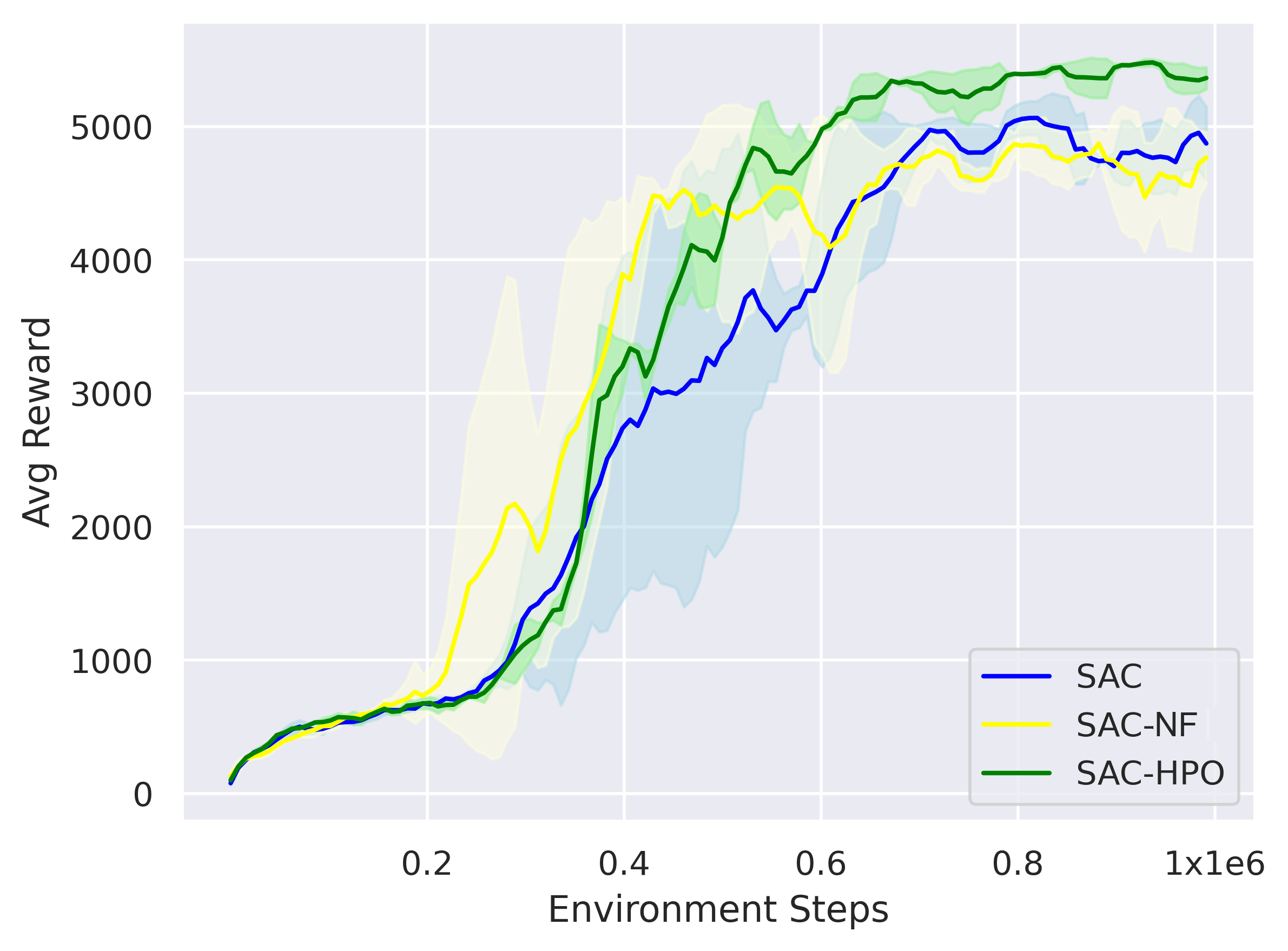}
    \label{fig:mujoco_5}
}
\subfigure[Humanoid PyBullet]{
    \includegraphics[width=1.5in]{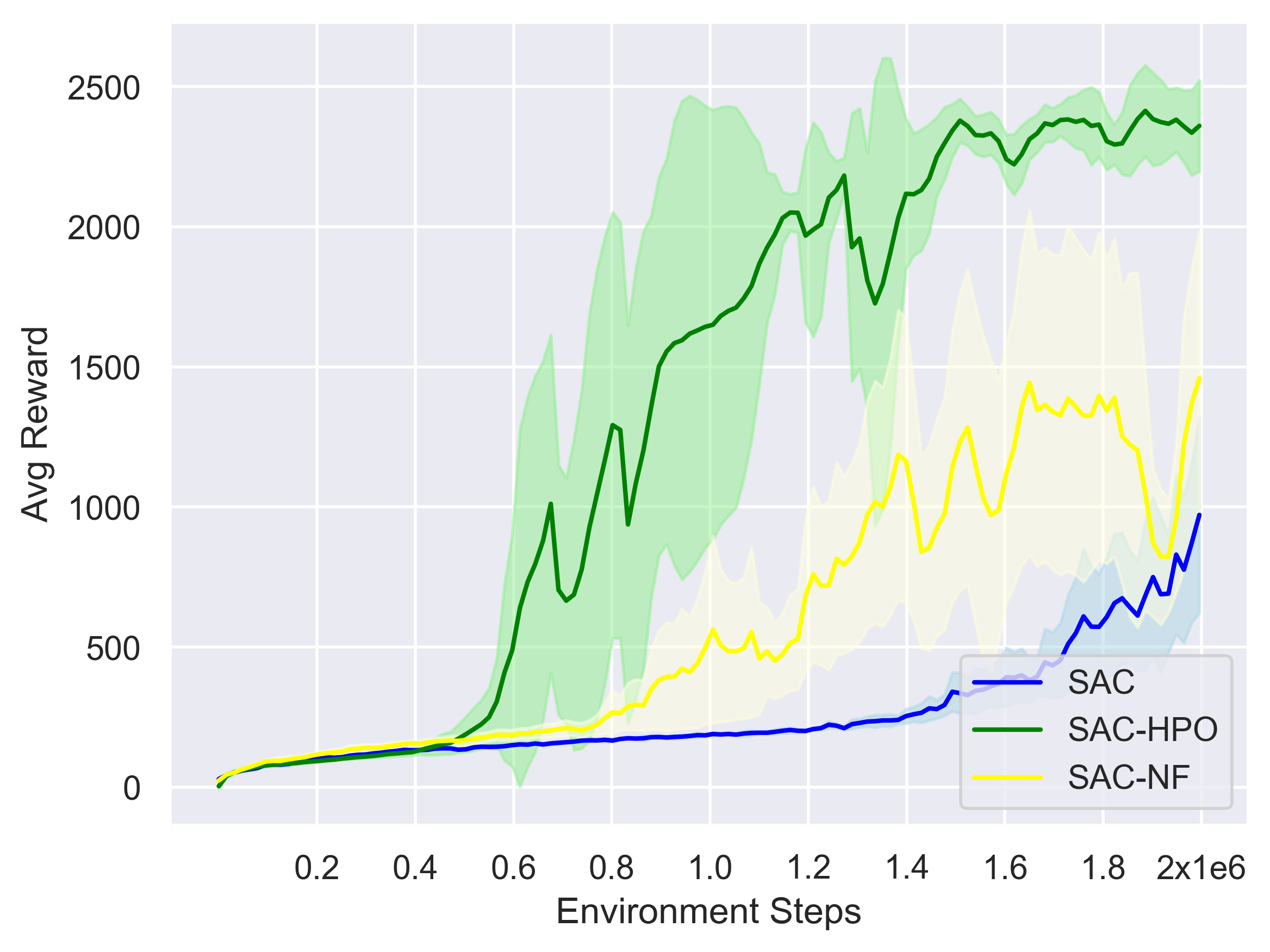}
    \label{fig:bullet_6}
}
\subfigure[Flagrun PyBullet]{
    \includegraphics[width=1.5in]{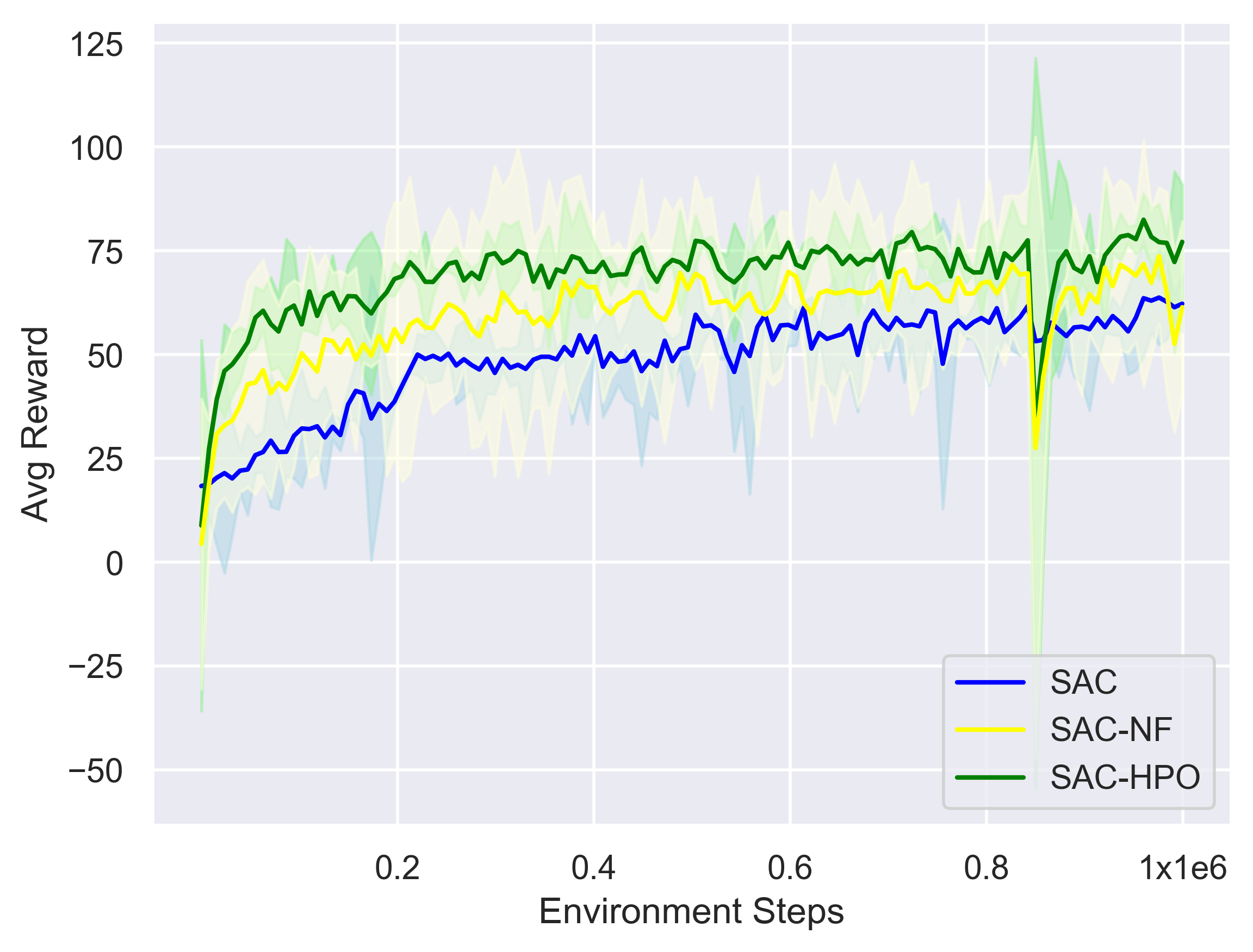}
    \label{fig:bullet_7}
}
\subfigure[Flagrun Harder PyBullet]{
    \includegraphics[width=1.5in]{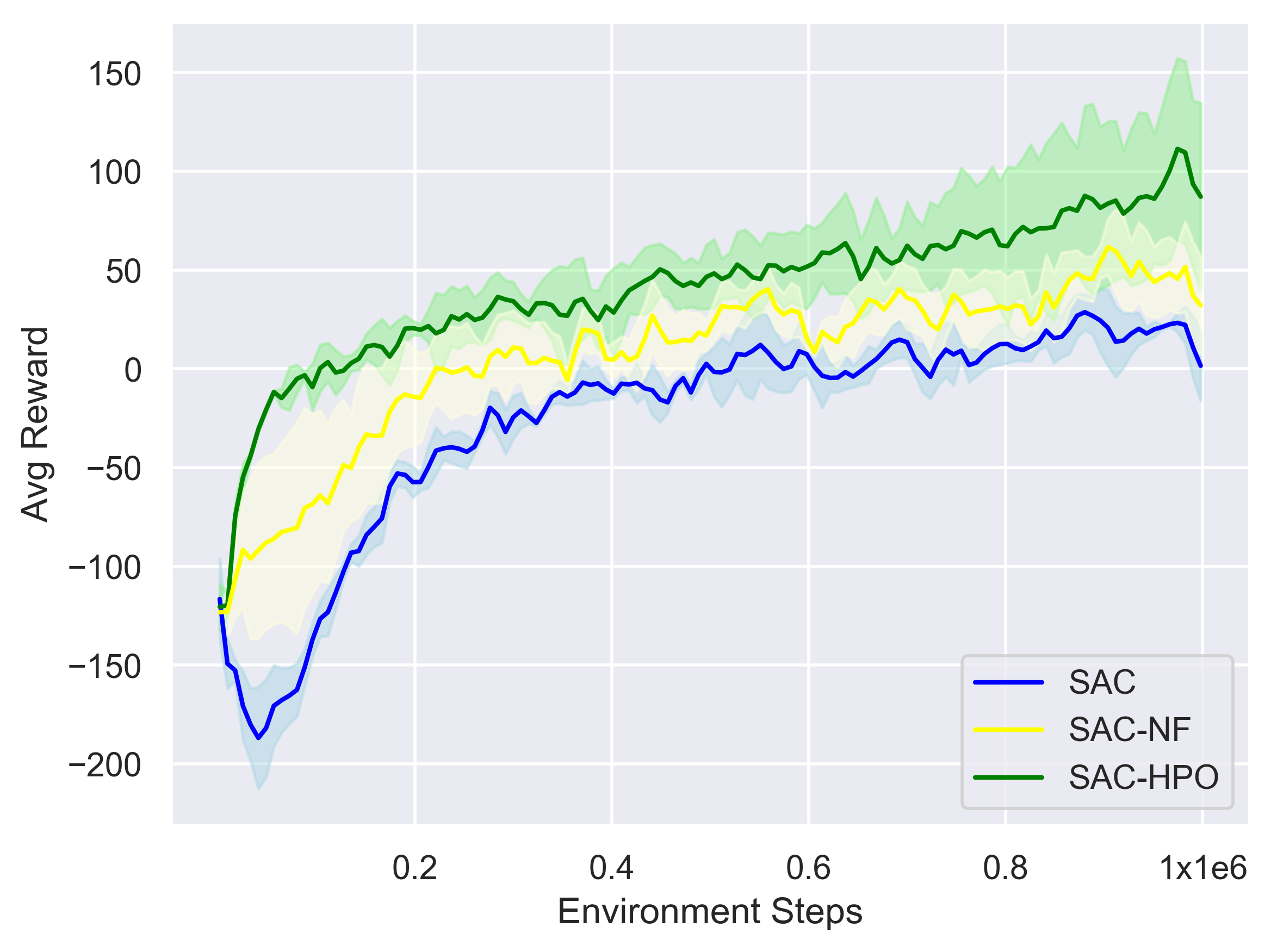}
    \label{fig:bullet_8}
}
\caption{The learning performance comparison over 8 tasks. All the curves are averaged over 5 random seeds, where shadowed regions are standard deviations.} 
\label{fig:mujoco}
\end{figure*}

\begin{algorithm}[ht]
\caption{Hamiltonian Policy in safe RL; $s, \pi_{\theta}, \beta_0, \epsilon, K, d_0$}
\label{alg:safe}
\begin{algorithmic}[1]
\State Sample $a_0\sim\pi_{\theta}(\cdot|s), \rho_0\sim\mathcal{N}(0, \beta_0^{-1}I)$
\For{$k=0,\ldots,K-1$}
\State Transform $(a_k, \rho_k)$ to $(a_{k+1}, \rho_{k+1})$ by the proposed leapfrog \eqref{leapfrog1} 
\If{$(a_{k+1}, \rho_{k+1})$ satisfies Lyapunov constraint \eqref{lyapunov}}
\State Return $a_{k+1}, \rho_{k+1}$
\EndIf
\State Sample $\tilde{\rho}\sim\mathcal{N}(0, \beta_0^{-1}I)$, and update $\rho_{k+1}\longleftarrow\rho_{k+1}+\tilde{\rho}$
\EndFor
\State Return {$a_K, \rho_K$}
\end{algorithmic}
\end{algorithm}

%% file: pol.pdf_tex
%% Accompanies image file 'pol.pdf' (pdf, eps, ps)
\begingroup%
  \makeatletter%
  \providecommand\color[2][]{%
    \errmessage{(Inkscape) Color is used for the text in Inkscape, but the package 'color.sty' is not loaded}%
    \renewcommand\color[2][]{}%
  }%
  \providecommand\transparent[1]{%
    \errmessage{(Inkscape) Transparency is used (non-zero) for the text in Inkscape, but the package 'transparent.sty' is not loaded}%
    \renewcommand\transparent[1]{}%
  }%
  \providecommand\rotatebox[2]{#2}%
  \newcommand*\fsize{\dimexpr\f@size pt\relax}%
  \newcommand*\lineheight[1]{\fontsize{\fsize}{#1\fsize}\selectfont}%
  \ifx\svgwidth\undefined%
    \setlength{\unitlength}{346.02056925bp}%
    \ifx\svgscale\undefined%
      \relax%
    \else%
      \setlength{\unitlength}{\unitlength * \real{\svgscale}}%
    \fi%
  \else%
    \setlength{\unitlength}{\svgwidth}%
  \fi%
  \global\let\svgwidth\undefined%
  \global\let\svgscale\undefined%
  \makeatother%
  \begin{picture}(1,0.28524899)%
    \lineheight{1}%
    \setlength\tabcolsep{0pt}%
    \put(-0.00218726,0.07204418){\color[rgb]{0,0,0}\makebox(0,0)[lt]{\lineheight{1.25}\smash{\begin{tabular}[t]{l}$s$\end{tabular}}}}%
    \put(0,0){\includegraphics[width=\unitlength,page=1]{pol.pdf}}%
    \put(0.32270868,0.05666331){\color[rgb]{0,0,0}\makebox(0,0)[lt]{\lineheight{1.25}\smash{\begin{tabular}[t]{l}$a_0\sim\mathcal{N}(\cdot|\mu(s),\Sigma(s))$\end{tabular}}}}%
    \put(0.3106583,0.26300951){\color[rgb]{0,0,0}\makebox(0,0)[lt]{\lineheight{1.25}\smash{\begin{tabular}[t]{l}$\rho_0\sim\mathcal{N}(0,\beta_0^{-1} I)$\end{tabular}}}}%
    \put(0,0){\includegraphics[width=\unitlength,page=2]{pol.pdf}}%
    \put(0.58776053,0.11058433){\color[rgb]{0,0,0}\makebox(0,0)[lt]{\lineheight{1.25}\smash{\begin{tabular}[t]{l}$a_1$\end{tabular}}}}%
    \put(0.58945275,0.19367353){\color[rgb]{0,0,0}\makebox(0,0)[lt]{\lineheight{1.25}\smash{\begin{tabular}[t]{l}$\rho_1$\end{tabular}}}}%
    \put(0,0){\includegraphics[width=\unitlength,page=3]{pol.pdf}}%
    \put(0.81259893,0.109722){\color[rgb]{0,0,0}\makebox(0,0)[lt]{\lineheight{1.25}\smash{\begin{tabular}[t]{l}$a_2$\end{tabular}}}}%
    \put(0.80995607,0.19714621){\color[rgb]{0,0,0}\makebox(0,0)[lt]{\lineheight{1.25}\smash{\begin{tabular}[t]{l}$\rho_2$\\\end{tabular}}}}%
    \put(0.8820856,0.15466341){\color[rgb]{0,0,0}\makebox(0,0)[lt]{\lineheight{1.25}\smash{\begin{tabular}[t]{l}$\ldots$\end{tabular}}}}%
    \put(0,0){\includegraphics[width=\unitlength,page=4]{pol.pdf}}%
    \put(0.67969156,0.14923888){\color[rgb]{0,0,0}\makebox(0,0)[lt]{\lineheight{1.25}\smash{\begin{tabular}[t]{l}HMC\end{tabular}}}}%
    \put(0,0){\includegraphics[width=\unitlength,page=5]{pol.pdf}}%
    \put(0.45283976,0.14917141){\color[rgb]{0,0,0}\makebox(0,0)[lt]{\lineheight{1.25}\smash{\begin{tabular}[t]{l}HMC\end{tabular}}}}%
    \put(0.14106841,0.05026477){\color[rgb]{0,0,0}\makebox(0,0)[lt]{\lineheight{1.25}\smash{\begin{tabular}[t]{l}$\pi_{\theta}$\end{tabular}}}}%
  \end{picture}%
\endgroup%

%% file: hmc.pdf_tex
%% Accompanies image file 'hmc.pdf' (pdf, eps, ps)
\begingroup%
  \makeatletter%
  \providecommand\color[2][]{%
    \errmessage{(Inkscape) Color is used for the text in Inkscape, but the package 'color.sty' is not loaded}%
    \renewcommand\color[2][]{}%
  }%
  \providecommand\transparent[1]{%
    \errmessage{(Inkscape) Transparency is used (non-zero) for the text in Inkscape, but the package 'transparent.sty' is not loaded}%
    \renewcommand\transparent[1]{}%
  }%
  \providecommand\rotatebox[2]{#2}%
  \newcommand*\fsize{\dimexpr\f@size pt\relax}%
  \newcommand*\lineheight[1]{\fontsize{\fsize}{#1\fsize}\selectfont}%
  \ifx\svgwidth\undefined%
    \setlength{\unitlength}{556.03600053bp}%
    \ifx\svgscale\undefined%
      \relax%
    \else%
      \setlength{\unitlength}{\unitlength * \real{\svgscale}}%
    \fi%
  \else%
    \setlength{\unitlength}{\svgwidth}%
  \fi%
  \global\let\svgwidth\undefined%
  \global\let\svgscale\undefined%
  \makeatother%
  \begin{picture}(1,0.48327169)%
    \lineheight{1}%
    \setlength\tabcolsep{0pt}%
    \put(0,0){\includegraphics[width=\unitlength,page=1]{hmc.pdf}}%
    \put(0.27176542,0.20817816){\color[rgb]{0,0,0}\makebox(0,0)[lt]{\lineheight{1.25}\smash{\begin{tabular}[t]{l}$T_h$\end{tabular}}}}%
    \put(0,0){\includegraphics[width=\unitlength,page=2]{hmc.pdf}}%
    \put(0.13788082,0.20944421){\color[rgb]{0,0,0}\makebox(0,0)[lt]{\lineheight{1.25}\smash{\begin{tabular}[t]{l}$\sigma_h$\end{tabular}}}}%
    \put(0,0){\includegraphics[width=\unitlength,page=3]{hmc.pdf}}%
    \put(0.15540307,0.11605845){\color[rgb]{0,0,0}\makebox(0,0)[lt]{\lineheight{1.25}\smash{\begin{tabular}[t]{l}$g$\end{tabular}}}}%
    \put(0,0){\includegraphics[width=\unitlength,page=4]{hmc.pdf}}%
    \put(0.75603354,0.20775518){\color[rgb]{0,0,0}\makebox(0,0)[lt]{\lineheight{1.25}\smash{\begin{tabular}[t]{l}$T_h$\end{tabular}}}}%
    \put(0,0){\includegraphics[width=\unitlength,page=5]{hmc.pdf}}%
    \put(0.62484673,0.20902123){\color[rgb]{0,0,0}\makebox(0,0)[lt]{\lineheight{1.25}\smash{\begin{tabular}[t]{l}$\sigma_h$\end{tabular}}}}%
    \put(0,0){\includegraphics[width=\unitlength,page=6]{hmc.pdf}}%
    \put(0.63967083,0.1156354){\color[rgb]{0,0,0}\makebox(0,0)[lt]{\lineheight{1.25}\smash{\begin{tabular}[t]{l}$g$\end{tabular}}}}%
    \put(0,0){\includegraphics[width=\unitlength,page=7]{hmc.pdf}}%
    \put(0.84625439,0.4395967){\color[rgb]{0,0,0}\makebox(0,0)[lt]{\lineheight{1.25}\smash{\begin{tabular}[t]{l}${\rho}_{k+1}$\end{tabular}}}}%
    \put(0,0){\includegraphics[width=\unitlength,page=8]{hmc.pdf}}%
    \put(0.28024997,0.27809918){\color[rgb]{0,0,0}\makebox(0,0)[lt]{\lineheight{1.25}\smash{\begin{tabular}[t]{l}$\times$\end{tabular}}}}%
    \put(0,0){\includegraphics[width=\unitlength,page=9]{hmc.pdf}}%
    \put(0.47185702,0.43929092){\color[rgb]{0,0,0}\makebox(0,0)[lt]{\lineheight{1.25}\smash{\begin{tabular}[t]{l}$\rho_{k+\frac{1}{2}}$\end{tabular}}}}%
    \put(0,0){\includegraphics[width=\unitlength,page=10]{hmc.pdf}}%
    \put(0.02397589,0.4397518){\color[rgb]{0,0,0}\makebox(0,0)[lt]{\lineheight{1.25}\smash{\begin{tabular}[t]{l}$\rho_k$\end{tabular}}}}%
    \put(0,0){\includegraphics[width=\unitlength,page=11]{hmc.pdf}}%
    \put(0.84664131,0.03358266){\color[rgb]{0,0,0}\makebox(0,0)[lt]{\lineheight{1.25}\smash{\begin{tabular}[t]{l}${a}_{k+1}$\end{tabular}}}}%
    \put(0,0){\includegraphics[width=\unitlength,page=12]{hmc.pdf}}%
    \put(0.47725236,0.03088499){\color[rgb]{0,0,0}\makebox(0,0)[lt]{\lineheight{1.25}\smash{\begin{tabular}[t]{l}$a_{k+1}$\end{tabular}}}}%
    \put(0,0){\includegraphics[width=\unitlength,page=13]{hmc.pdf}}%
    \put(0.02127823,0.03088566){\color[rgb]{0,0,0}\makebox(0,0)[lt]{\lineheight{1.25}\smash{\begin{tabular}[t]{l}$a_k$\end{tabular}}}}%
    \put(0,0){\includegraphics[width=\unitlength,page=14]{hmc.pdf}}%
    \put(0.14312272,0.27587361){\color[rgb]{0,0,0}\makebox(0,0)[lt]{\lineheight{1.25}\smash{\begin{tabular}[t]{l}$1-$\end{tabular}}}}%
    \put(0,0){\includegraphics[width=\unitlength,page=15]{hmc.pdf}}%
    \put(0.01670429,0.27813057){\color[rgb]{0,0,0}\makebox(0,0)[lt]{\lineheight{1.25}\smash{\begin{tabular}[t]{l}$\times$\end{tabular}}}}%
    \put(0,0){\includegraphics[width=\unitlength,page=16]{hmc.pdf}}%
    \put(0.14536124,0.35536811){\color[rgb]{0,0,0}\makebox(0,0)[lt]{\lineheight{1.25}\smash{\begin{tabular}[t]{l}$+$\end{tabular}}}}%
    \put(0,0){\includegraphics[width=\unitlength,page=17]{hmc.pdf}}%
    \put(0.27383059,0.35673933){\color[rgb]{0,0,0}\makebox(0,0)[lt]{\lineheight{1.25}\smash{\begin{tabular}[t]{l}$\odot\frac{\epsilon}{2}$\end{tabular}}}}%
    \put(0,0){\includegraphics[width=\unitlength,page=18]{hmc.pdf}}%
    \put(0.28192375,0.4343937){\color[rgb]{0,0,0}\makebox(0,0)[lt]{\lineheight{1.25}\smash{\begin{tabular}[t]{l}$+$\end{tabular}}}}%
    \put(0,0){\includegraphics[width=\unitlength,page=19]{hmc.pdf}}%
    \put(0.37980942,0.20991946){\color[rgb]{0,0,0}\makebox(0,0)[lt]{\lineheight{1.25}\smash{\begin{tabular}[t]{l}$\odot\epsilon$\end{tabular}}}}%
    \put(0,0){\includegraphics[width=\unitlength,page=20]{hmc.pdf}}%
    \put(0.38756526,0.02311671){\color[rgb]{0,0,0}\makebox(0,0)[lt]{\lineheight{1.25}\smash{\begin{tabular}[t]{l}$+$\end{tabular}}}}%
    \put(0,0){\includegraphics[width=\unitlength,page=21]{hmc.pdf}}%
    \put(0.5063675,0.27500988){\color[rgb]{0,0,0}\makebox(0,0)[lt]{\lineheight{1.25}\smash{\begin{tabular}[t]{l}$\times$\end{tabular}}}}%
    \put(0,0){\includegraphics[width=\unitlength,page=22]{hmc.pdf}}%
    \put(0.62739104,0.27275296){\color[rgb]{0,0,0}\makebox(0,0)[lt]{\lineheight{1.25}\smash{\begin{tabular}[t]{l}$1-$\end{tabular}}}}%
    \put(0,0){\includegraphics[width=\unitlength,page=23]{hmc.pdf}}%
    \put(0.62693185,0.35494513){\color[rgb]{0,0,0}\makebox(0,0)[lt]{\lineheight{1.25}\smash{\begin{tabular}[t]{l}$+$\end{tabular}}}}%
    \put(0,0){\includegraphics[width=\unitlength,page=24]{hmc.pdf}}%
    \put(0.75540104,0.35901402){\color[rgb]{0,0,0}\makebox(0,0)[lt]{\lineheight{1.25}\smash{\begin{tabular}[t]{l}$\odot\frac{\epsilon}{2}$\end{tabular}}}}%
    \put(0,0){\includegraphics[width=\unitlength,page=25]{hmc.pdf}}%
    \put(0.76721573,0.2776762){\color[rgb]{0,0,0}\makebox(0,0)[lt]{\lineheight{1.25}\smash{\begin{tabular}[t]{l}$\times$\end{tabular}}}}%
    \put(0,0){\includegraphics[width=\unitlength,page=26]{hmc.pdf}}%
    \put(0.76480094,0.4397737){\color[rgb]{0,0,0}\makebox(0,0)[lt]{\lineheight{1.25}\smash{\begin{tabular}[t]{l}$+$\end{tabular}}}}%
  \end{picture}%
\endgroup%

%% file: part3.tex
\section{Experiment}
In experiments, the Hamiltonian policy is applied into soft actor critic (SAC), so the proposed method is denoted as "SAC-HPO". The environments in our experiments are diverse, ranging from OpenAI Gym MuJoCo \cite{todorov2012mujoco} to the realistic Roboschool PyBullet suit \cite{coumans2016pybullet,benelot2018}. We empirically evaluate the proposed method from many perspectives. First, SAC-HPO is compared with the primitive SAC \cite{haarnoja2018soft} and SAC-NF \cite{mazoure2020leveraging} to show our advantage over classical SAC and normalizing flow policy. Second, we show the advantage of Hamiltonian policy in two MuJoCo environments with safety constraints, comparing with SAC-Lagrangian \cite{tessler2018reward,chow2019lyapunov}.

In addition, we conduct ablation study on the proposed leapfrog operator and the sensitivity analysis of hyper-parameters in Section \ref{sec:analysis}. %By analyzing the effect of random momentum variable in Appendix \ref{sec:nonoise}, we did the comparison against iterative amortization policy optimization \cite{marino2020iterative}.
The effect of random momentum variables in HMC is also evaluated in Section \ref{sec:nonoise}.
Finally, in Section \ref{sec:shape}, we also analyze the shape of action distribution after leapfrog steps, verifying its non-Gaussianity and improvement of expressivity.

\begin{table*}[ht]
    \centering
    \begin{tabular}{p{0.9in}<{\centering}|p{0.2in}<{\centering}|p{0.35in}<{\centering}|p{0.35in}<{\centering}|p{0.3in}<{\centering}|p{0.2in}<{\centering}|p{0.2in}<{\centering}|p{0.2in}<{\centering}|p{0.2in}<{\centering}|p{0.3in}<{\centering}}
    \hline
         & $\alpha$ & $1/\sqrt{\beta_0^{\text{tr}}}$ & $1/\sqrt{\beta_0^{\text{exp}}}$ & $\epsilon$ & $K$ & $l$ & $h_n$ & $m$ & $\mathcal{B}$ size  \\\hline
      HalfCheetah-v2 & $0.2$ & $1$ & $0.5$ & $0.2$ & 3 & 1 & 32 & \multirow{8}*{256} & \multirow{8}*{$10^6$} \\
      Hopper-v2 & $0.2$ & $0.1$ & $1.5$ & $0.15$ & 2 & 1 & 32 &  &  \\
      Walker2d-v2 & $0.2$ & $0.2$ & $1.5$ & $0.15$ & 3 & 1 & 32 &  &  \\
      Ant-v2 & $0.2$ & $0.1$ & $1.0$ & $0.1$ & 3 & 1 & 32 &  &  \\
      Humanoid-v2 & $0.05$ & $1$ & $1$ & $0.1$ & 3 & 1 & 64 &  &  \\
      {Humanoid PyB.}  & $0.05$ & $0.4$ & $1.5$ & $0.2$ & 3 & 1 & 64 & & \\
      Flagrun  & $0.05$ & $0.2$ & $1$ & $0.15$ & 3 & 1 & 32 & & \\
      Flagrun Harder  & $0.05$ & $0.2$ & $1.5$ & $0.15$ & 3 & 1 & 64 & & \\ \hline
    \end{tabular}
    \makeatletter\def\@captype{table}\makeatother\caption{Hyperparameters in SAC-HPO.}
    \label{tab:hpo_hyp}
\end{table*}

\subsection{Continuous Control Tasks}
\label{sec:mujoco}

We compare SAC-HPO with SAC and SAC-NF on eight continuous control tasks. SAC is chosen because it is a fundamental learning method in actor-critic RL. SAC-NF is selected since it is a representative and widely-used method which adopts normalizing flow policy to improve the exploration. We use the official implementation of SAC \cite{haarnoja2018soft}. And we try our best to implement SAC-NF according to \cite{mazoure2020leveraging}, where the policy network is one-layer MLP with 256 hidden units and ReLU activation and radial normalizing flow is adopted.
The learning curves are shown in Figure \ref{fig:mujoco}, where first five tasks, corresponding from Figure \ref{fig:mujoco_1} to Figure \ref{fig:mujoco_5}, are from the MuJoCo suite and the other three are from Roboschool PyBullet. 
%We adopt the classical architecture for SAC and SAC-EBP \cite{haarnoja2017reinforcement,haarnoja2018soft}. 

\begin{figure*}[ht]
\centering
\subfigure[Ant-Reward]{
    \includegraphics[width=1.5in]{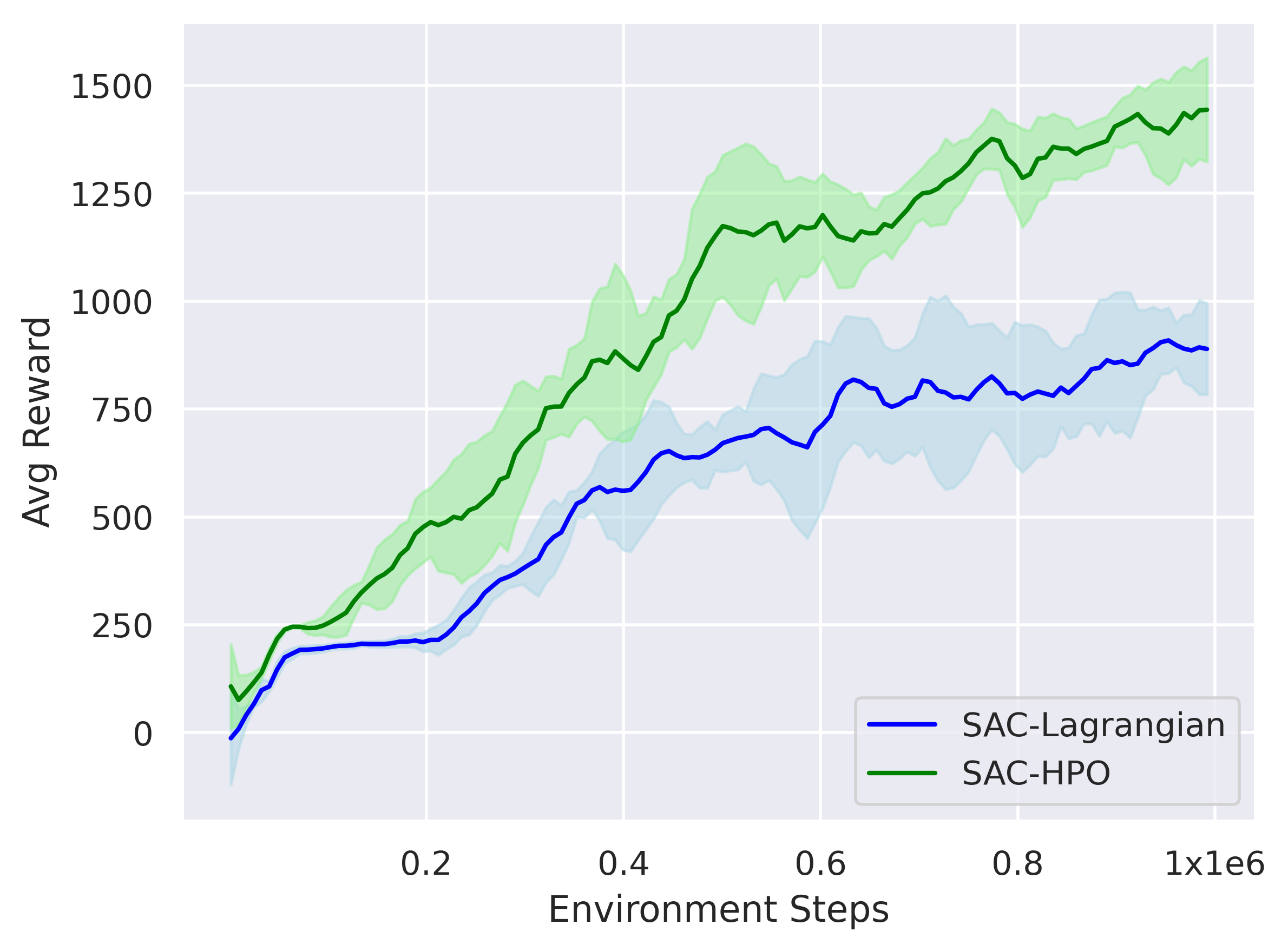}
    \label{fig:safe_ant}
}
\subfigure[Ant-Cost]{
    \includegraphics[width=1.5in]{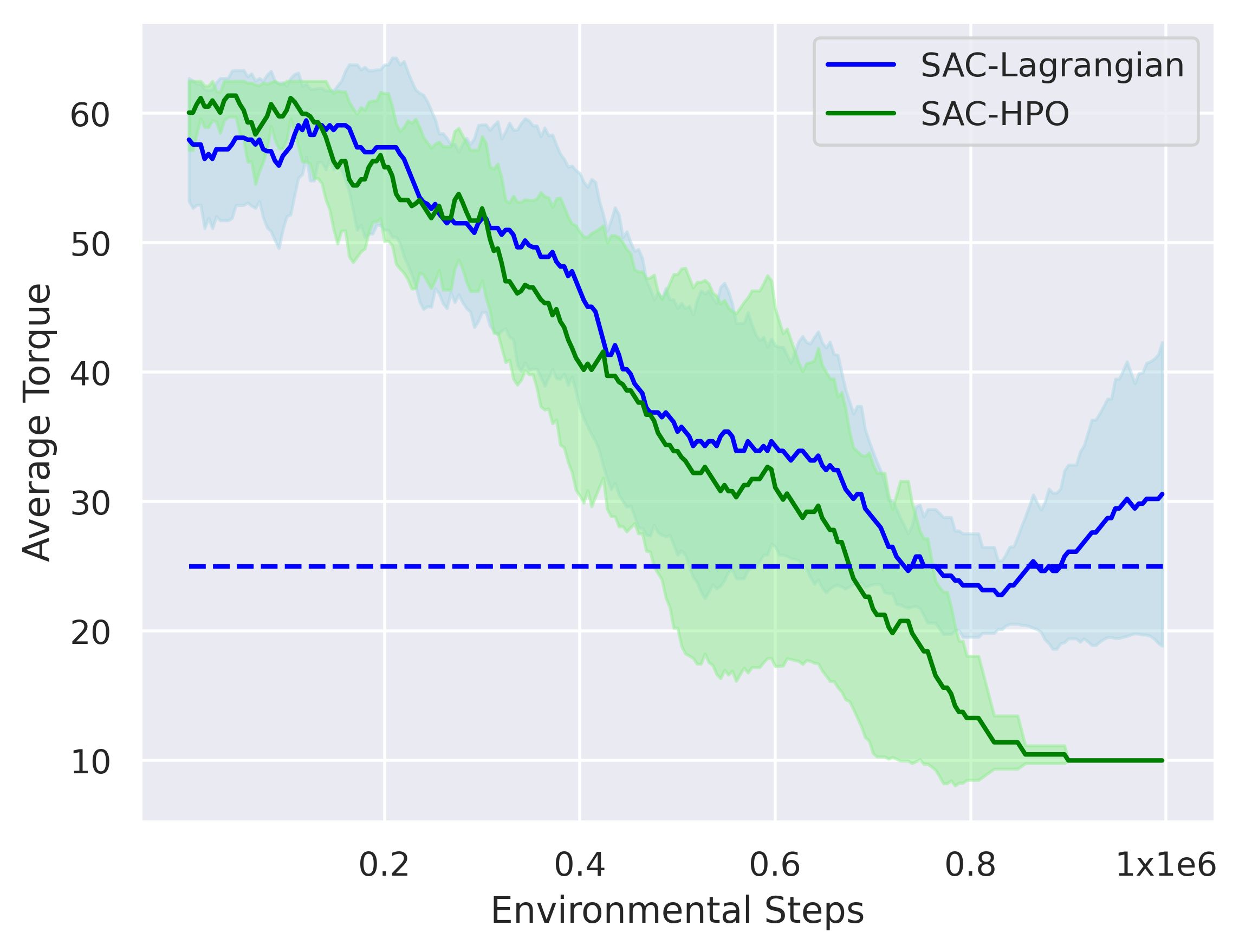}
    \label{fig:cost_ant}
}
\subfigure[HalfCheetah-Reward]{
    \includegraphics[width=1.5in]{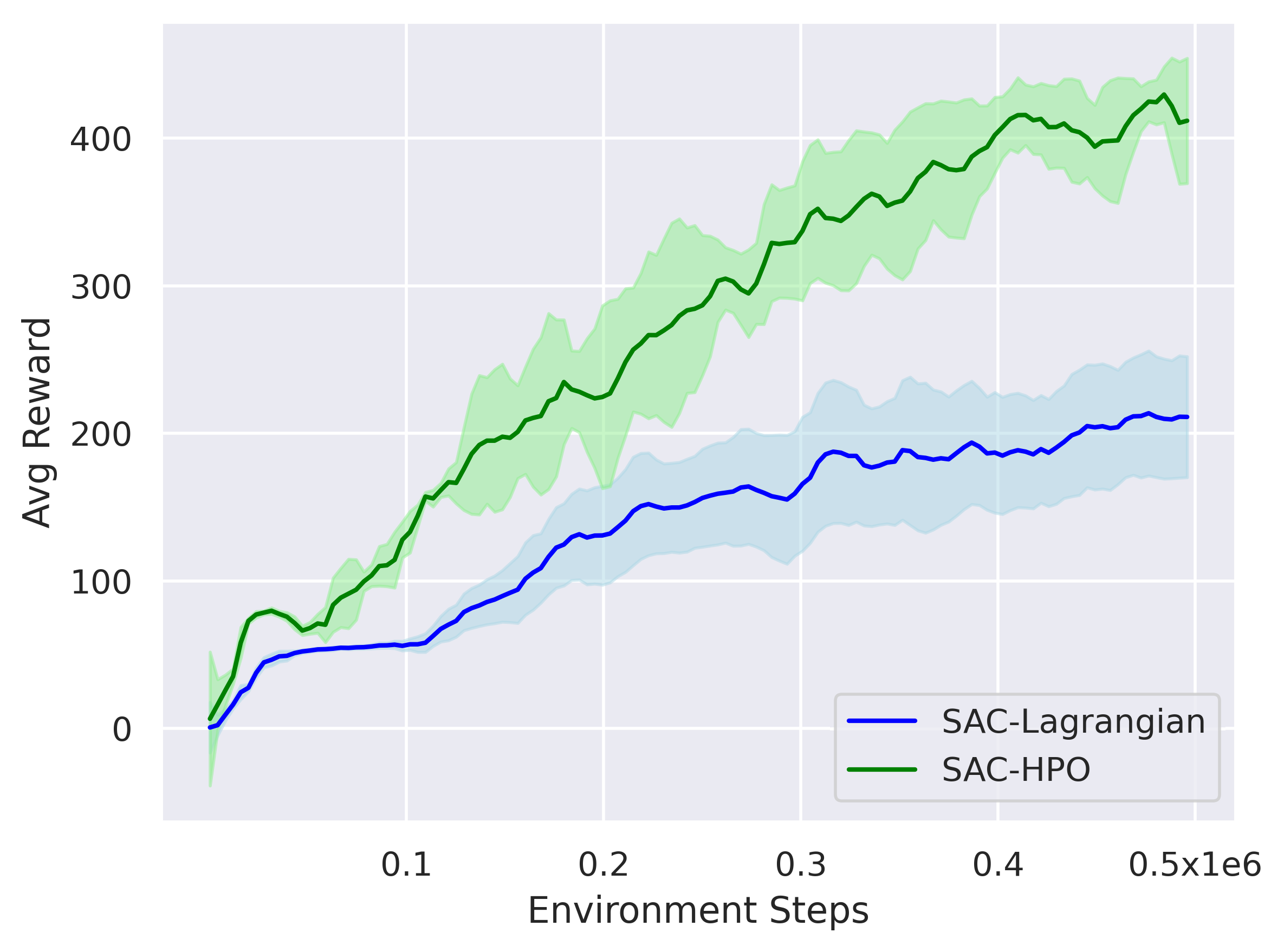}
    \label{fig:safe_halfcheetah}
}
\subfigure[HalfCheetah-Cost]{
    \includegraphics[width=1.5in]{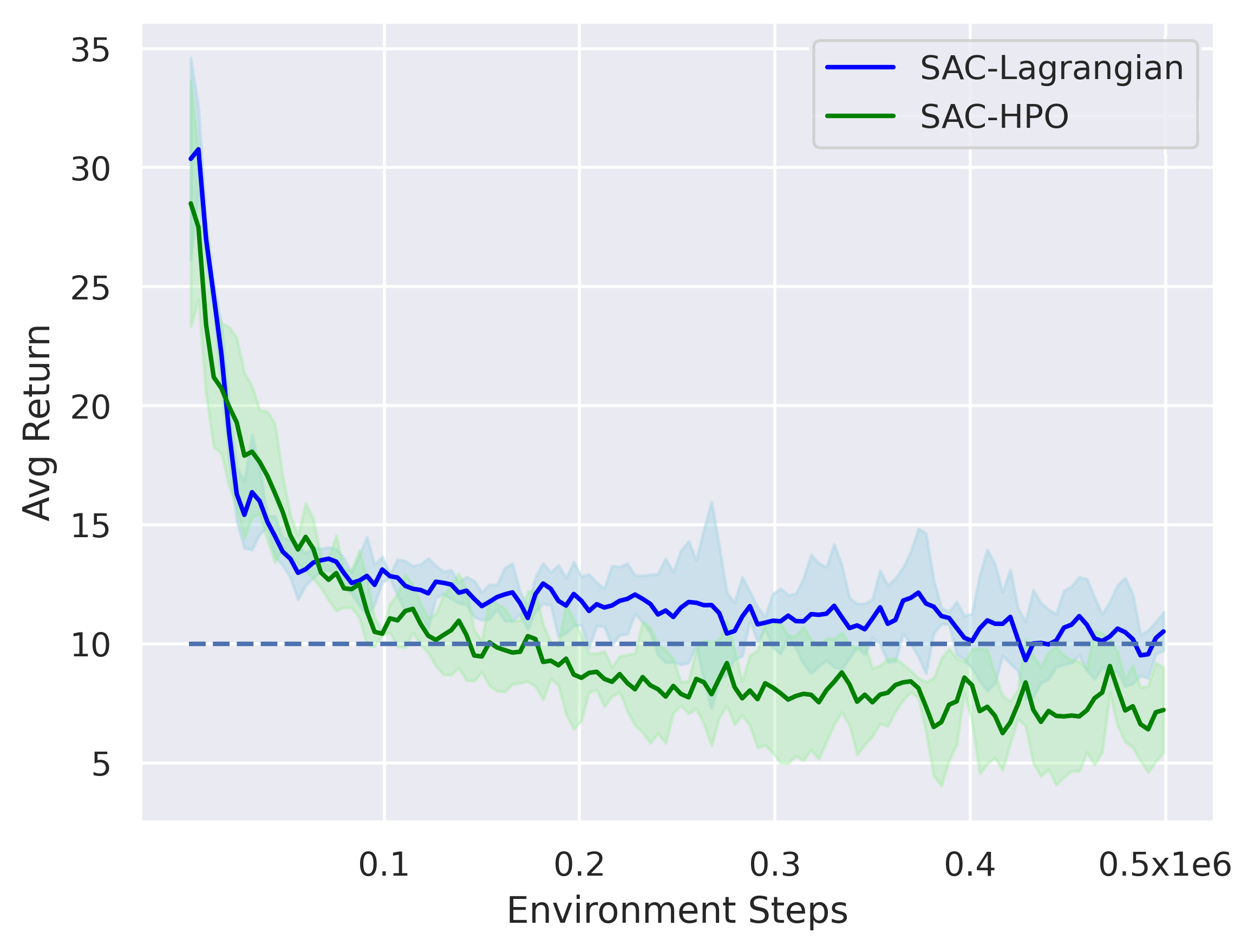}
    \label{fig:cost_halfcheetah}
}
\caption{Learning performance of Hamiltonian policy in safe RL.}
\label{fig:saferl}
\end{figure*}

All the methods use the same architecture for Q networks, hyper-parameters, and tuning scheme for the temperature $\alpha$. The critic (Q) networks follows the same architecture as \cite{haarnoja2018soft}, i.e., two-layer fully-connected neural networks with 256 units and ReLU activation in each layer, where two Q networks are implemented and trained by bootstrapping. All networks are updated by Adam optimizer \cite{kingma2014adam} with the learning rate of 3e-4. The batch size for updating policies and critics is 256, and the size of replay buffer is $10^6$.
In SAC, the policy network consists of two fully-connected hidden layers with 256 units and ReLU activation.

\begin{figure*}[ht]
\centering
\subfigure[Ant-v2]{
    \includegraphics[width=1.5in]{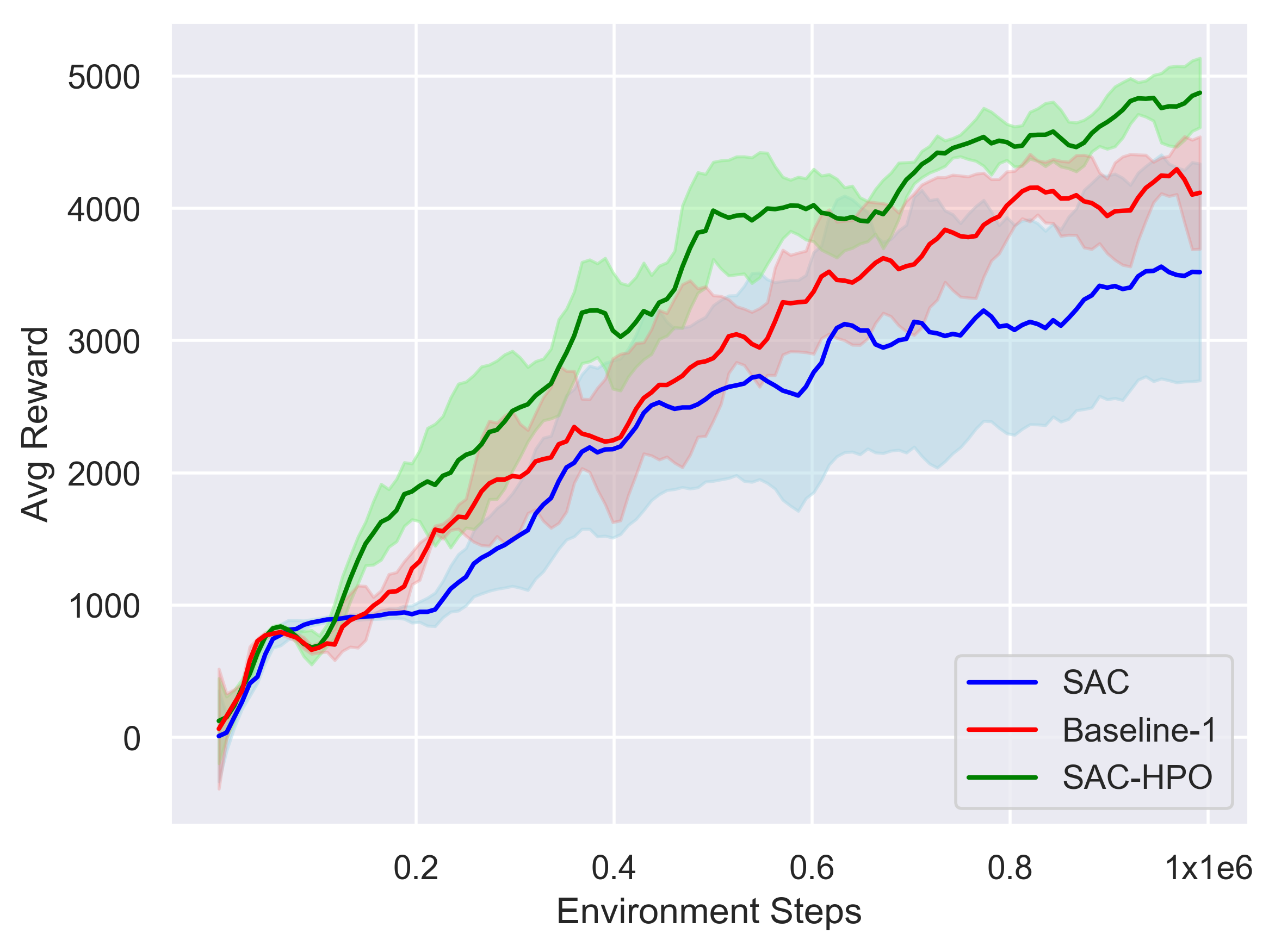}
    \label{fig:ablation_1}
}
\subfigure[Ant-v2]{
    \includegraphics[width=1.5in]{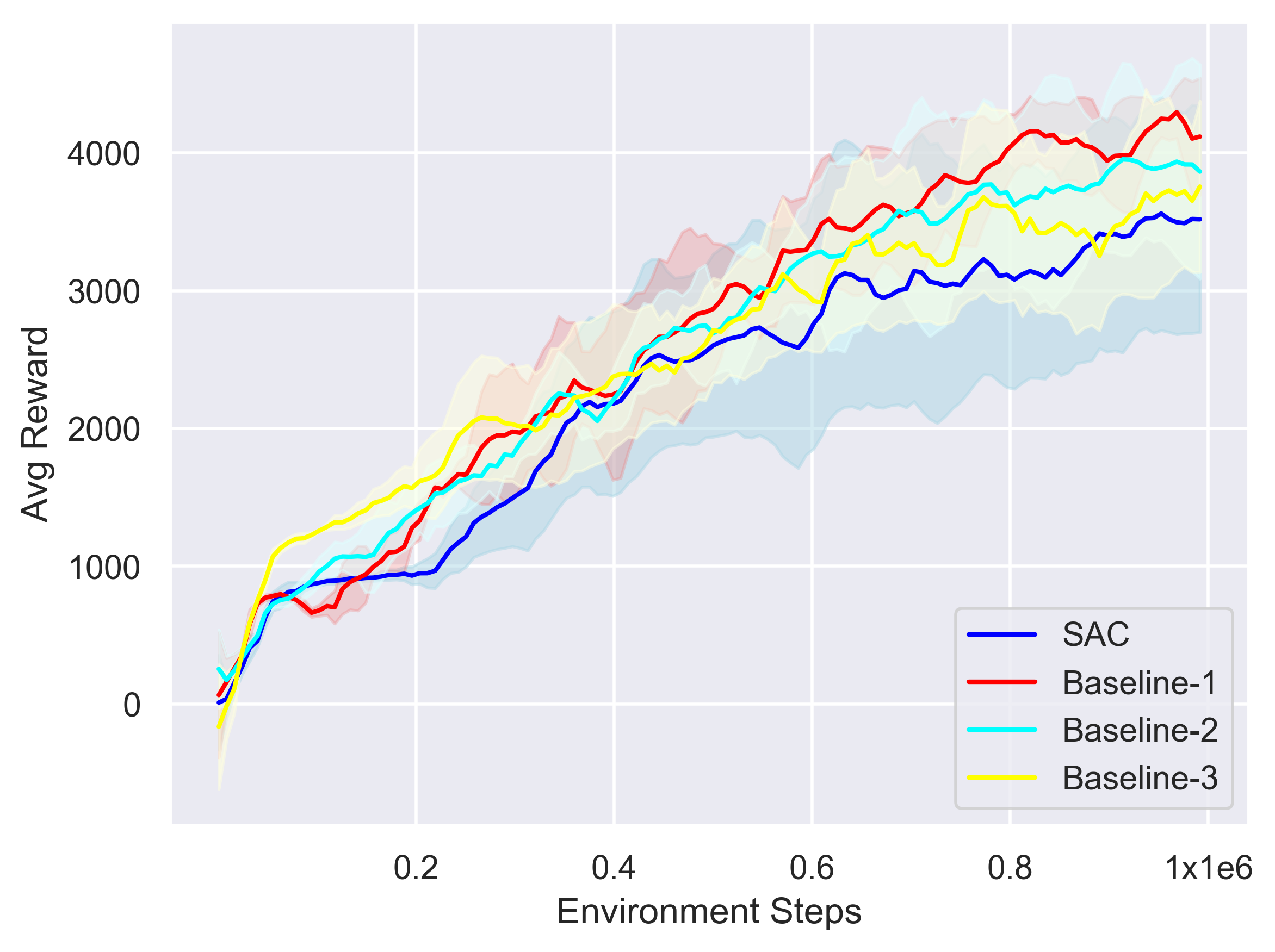}
    \label{fig:ablation_2}
}
\subfigure[Humanoid PyBullet]{
    \includegraphics[width=1.5in]{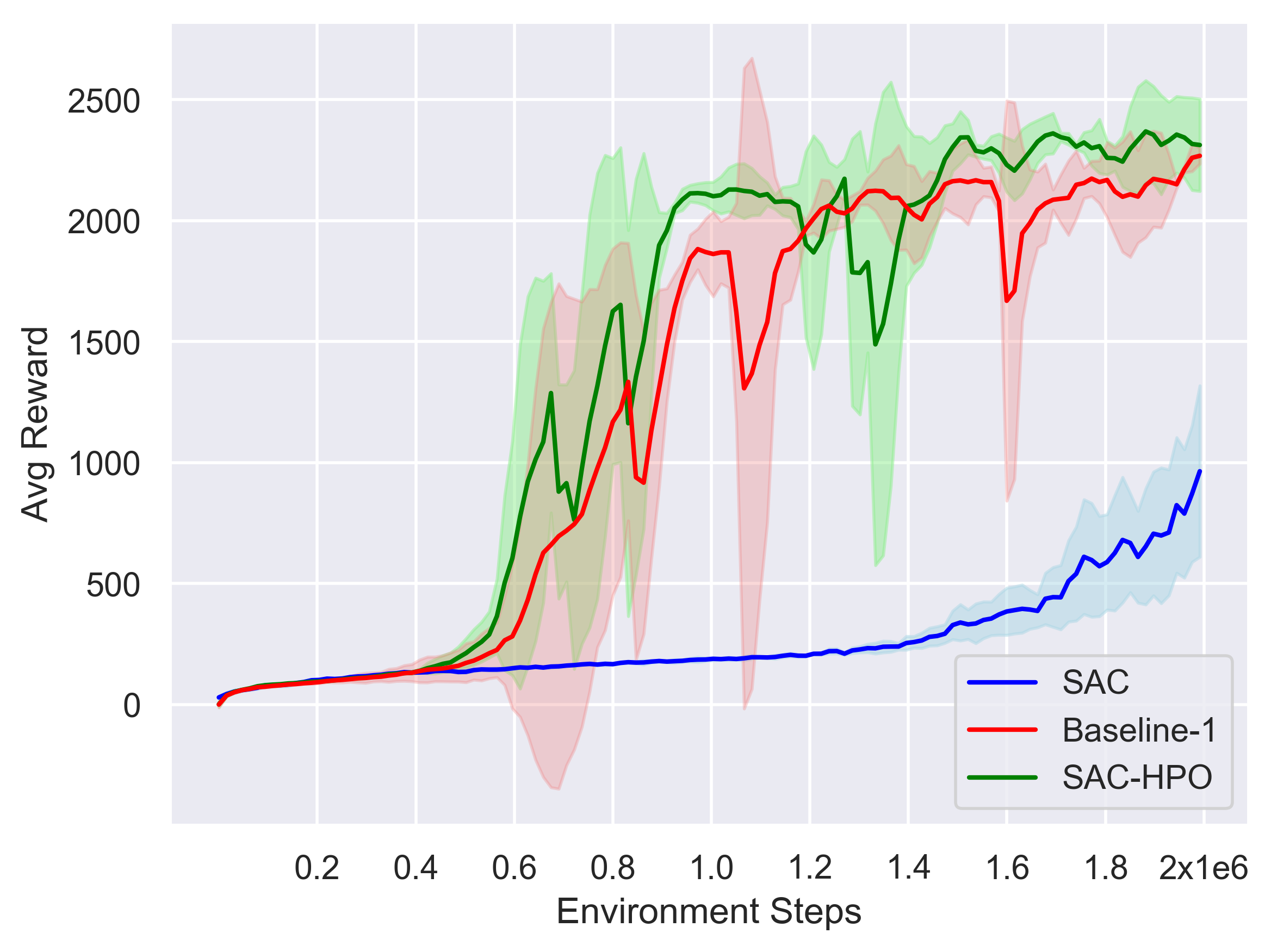}
    \label{fig:ablation_3}
}
\subfigure[Humanoid PyBullet]{
    \includegraphics[width=1.5in]{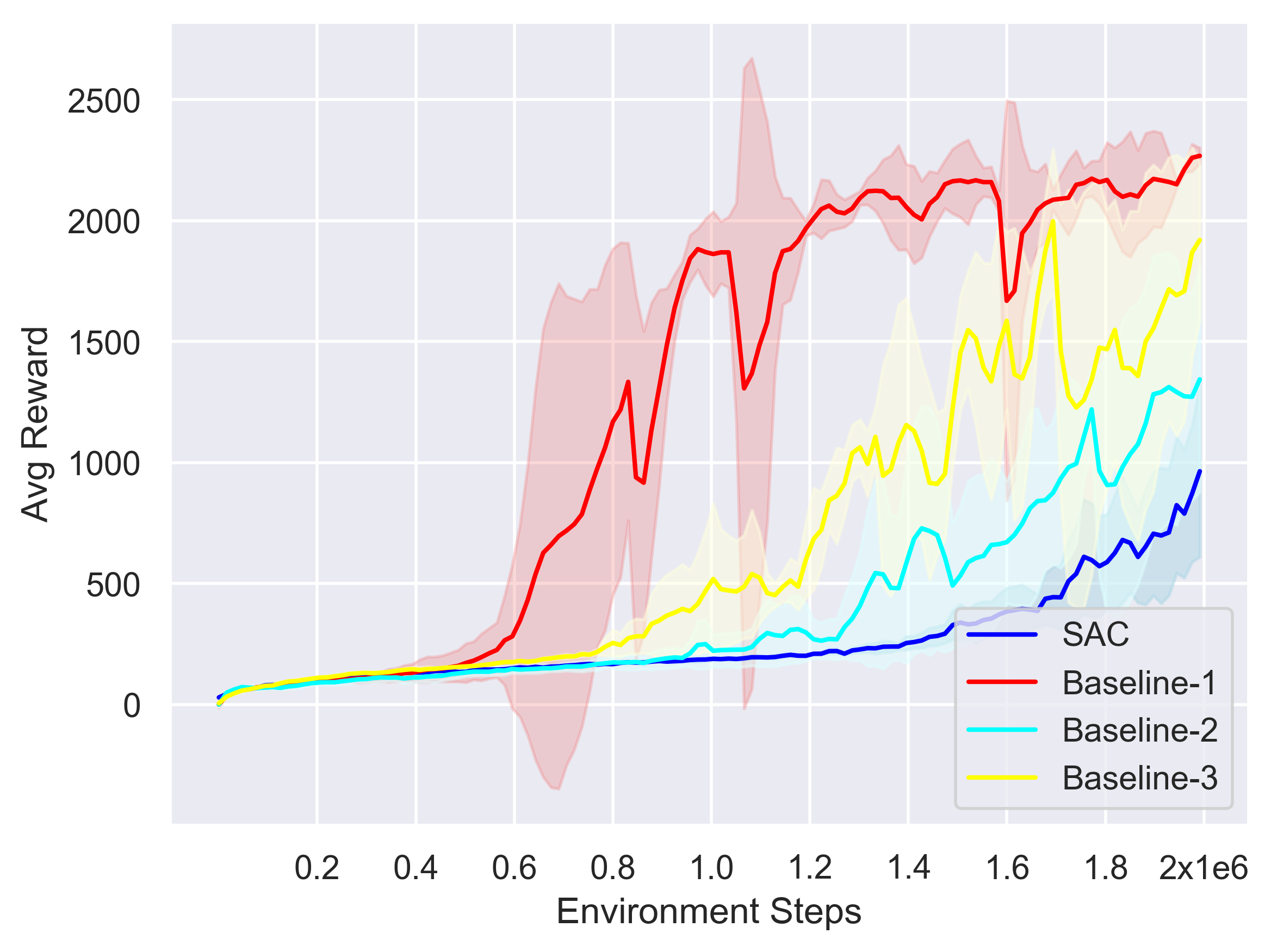}
    \label{fig:ablation_4}
}
\caption{Performance Comparison in Ablation Study.}
\label{fig:ablation}
\end{figure*}

In SAC-HPO, actions are evolved by HD simulated by leapfrog operations \eqref{leapfrog1} for $K\in\{1,2,3\}$ steps. The base policy only has one hidden layer with 256 units and ReLU activation. Neural networks in proposed leapfrog ($T_h$ and $\sigma_h$) are simple MLPs having one hidden layer with $h_n$ hidden units and ELU activation. The variances of the momentum vector ($\beta_0$) should be different for training and exploration, denoted as $\beta_0^{\text{tr}}$ and $\beta_0^{\text{exp}}$ respectively. In most experiments, we find the variance of momentum $\rho_0$ in exploration should be larger than that in training, i.e., $\beta_0^{\text{tr}}<\beta_0^{\text{exp}}$, which can improve exploration efficiency. Besides, it is important to make networks $T_h$ and $\sigma_h$ in leapfrog small, which can stabilize the learning process.  Hyper-parameters used in SAC-HPO are shown in Table \ref{tab:hpo_hyp}, including number of leapfrog steps ($K$), number of hidden units in $T_h$ and $\sigma_h$ ($h_n$), momentum variances for training and exploration ($\beta_0^{\text{tr}}$ and $\beta_0^{\text{exp}}$), update rate in leapfrog ($\epsilon$), and temperature parameter ($\alpha$). 

All results in Figure \ref{fig:mujoco} show the evaluation performance. Evaluation happens every 10,000 environmental steps, where each evaluation score (accumulated rewards in one episode) is averaged over 10 runs. The values reported in the plots are smoothed by exponential moving averaging (EMA) with a window size of 5, equivalent to averaging every 50,000 steps to improve comparability. 
We can see that the SAC-HPO outperforms SAC and SAC-NF in terms of both convergence rate and performance.

\subsection{Safe Reinforcement Learning}
\label{sec:safe}
In this section, we evaluate the performance of Hamiltonian policy in safe RL problems. The environments in this section are Ant-v2 and HalfCheetah-v2 in MuJoCo suite. At each step the robot selects the amount of torque to apply to each joint. In Ant-v2, the safety constraint is on the amount of torque the robot decided to apply at each time step \cite{tessler2018reward}. Since the target of the task is to prolong the motor life of the robot, the robot is constrained from using high torque values. This is accomplished by defining the constraint $d_0$ as the average torque the agent has applied to each motor. The constraint threshold on torques is set to be 25 in each episode. In HalfCheetah-v2, the safety constraint is that the speed of the robot should be less than 1, i.e., the constraint cost is $\bm{1}[|v|>1]$ at each time step \cite{chow2019lyapunov}. The constraint threshold $d_0$ is on the discounted sum of safety costs in each episode, which is set to be $10$.

In experiments, we learn critic networks for both return and accumulated costs (safety violations), denoted as $Q(s,a)$ and $Q_C(s,a)$ respectively. Both $Q$ and $Q_C$ are realized by two-layer MLP with 256 hidden units and ReLU activation in each layer. 

The baseline is the Lagrangian-based SAC \cite{chow2019lyapunov} which introduces a Lagrangian multiplier $\lambda$ to balance between return and safety costs, shorted as SAC-Lagrangian. The policy learning objective is $\max_{\theta}\mathbb{E}_{s\sim\mathcal{B},a\sim\pi_{\theta}(\cdot|s)}[(Q(s,a)-\lambda Q_C(s,a))-\alpha\log\pi_{\theta}(a|s)]$. And the multiplier is updated as $\lambda\longleftarrow[\lambda+\eta(J^{\pi_{\theta}}_C-d_0)]_+$ where $\eta=0.1$. Specifically, in Ant-v2, $J^{\pi_{\theta}}_C$ is the average sum of safety costs averaged in recent episodes, while in HalfCheetah-v2, $J^{\pi_{\theta}}_C$ is the discounted sum of safety costs averaged in recent episodes. 

In SAC-HPO, the Lyapunov constraint is written as \eqref{lyapunov}, where reference policy $\pi_B$ is the policy updated in the last iteration, and $Q_{\pi_B}, Q_{C,\pi_B}$ are target value networks of return and safety costs which are periodically updated in typical actor-critic RL algorithms. The working process of Hamiltonian policy in safe RL is summarized in Algorithm \ref{alg:safe} in Section \ref{sec:saferl}. In SAC-HPO, the policy learning objective is in the same form as \eqref{hmcobj}, where $Q_{\pi_{\theta}}$ is replaced by $Q_{\pi_B}+\lambda Q_{C, \pi_B}$ and $\lambda$ is updated in the same way as SAC-Lagrangian. The policy network and hyper-parameters of the SAC-HPO are same as Section \ref{sec:mujoco}, except that the maximum number of leapfrog $K$ is set to be $10$ which is larger than that in regular RL problems, and in practice the number of leapfrog steps taken actually is usually much smaller than $K$. 

The performance comparison is presented in Figure \ref{fig:saferl}, showing that our method not only improves the average return, but also reduces the safety violations. In safe RL, the learning objective contains $Q+\lambda Q_C$ instead of $Q$, where $\lambda$ is changing in every learning iteration. Hence, since the target posterior of actions in \eqref{target} is defined in terms of $Q+\lambda Q_C$, the amortization gap in policy optimization is more significant than that in regular RL due to the rapid changes of $\lambda$. So HMC is more necessary here to make the sampled actions better approximate the target posterior. Moreover, iterative HMC sampling can discard potentially unsafe actions until safe actions are sampled. So, Hamiltonian policy can achieve more significant performance improvement in safe RL tasks than that in regular RL tasks.

\begin{figure*}[ht]
\centering
\subfigure[Hopper-v2]{
    \includegraphics[width=1.5in]{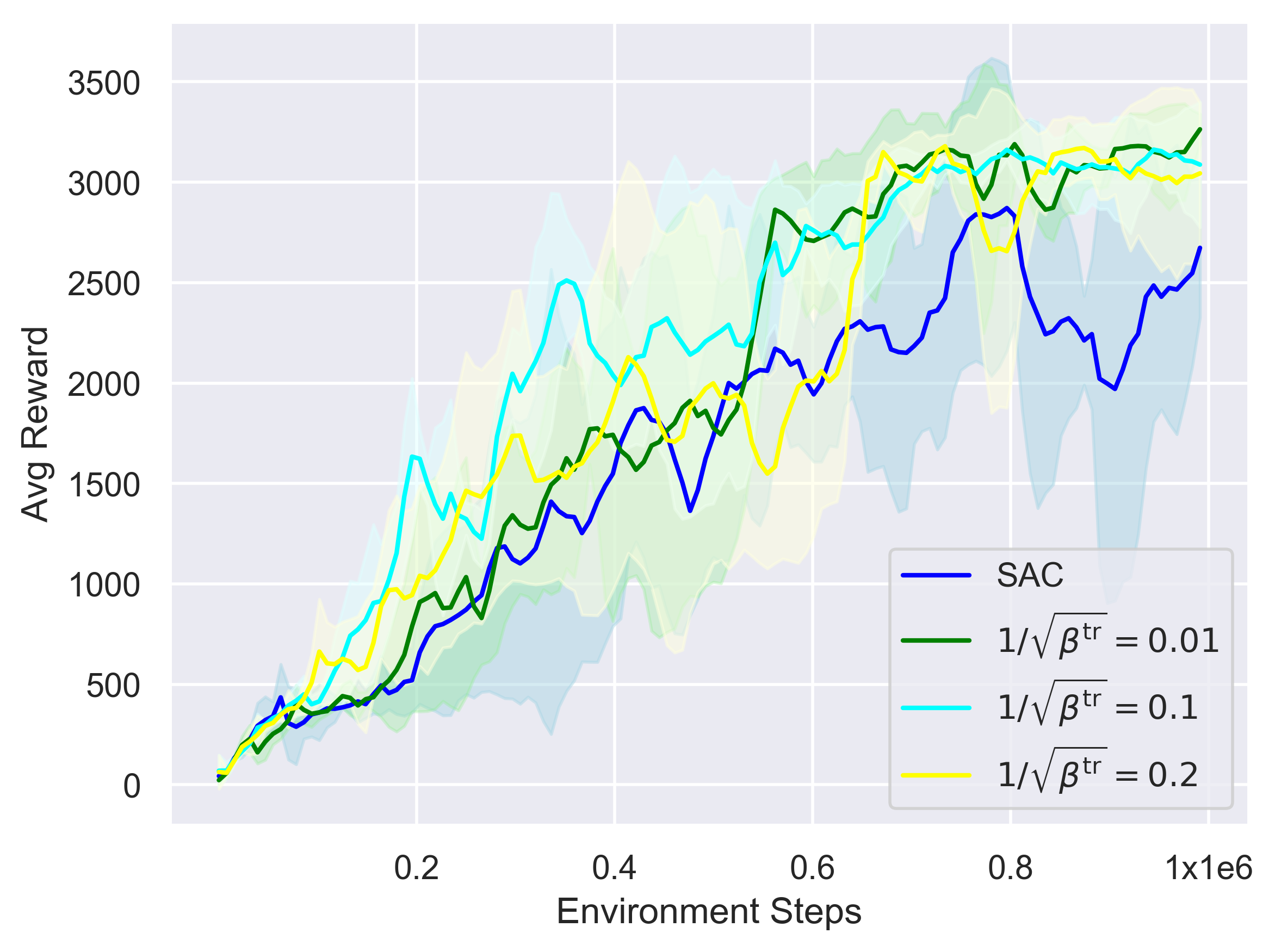}
    \label{fig:sens_1}
}
\subfigure[Ant-v2]{
    \includegraphics[width=1.5in]{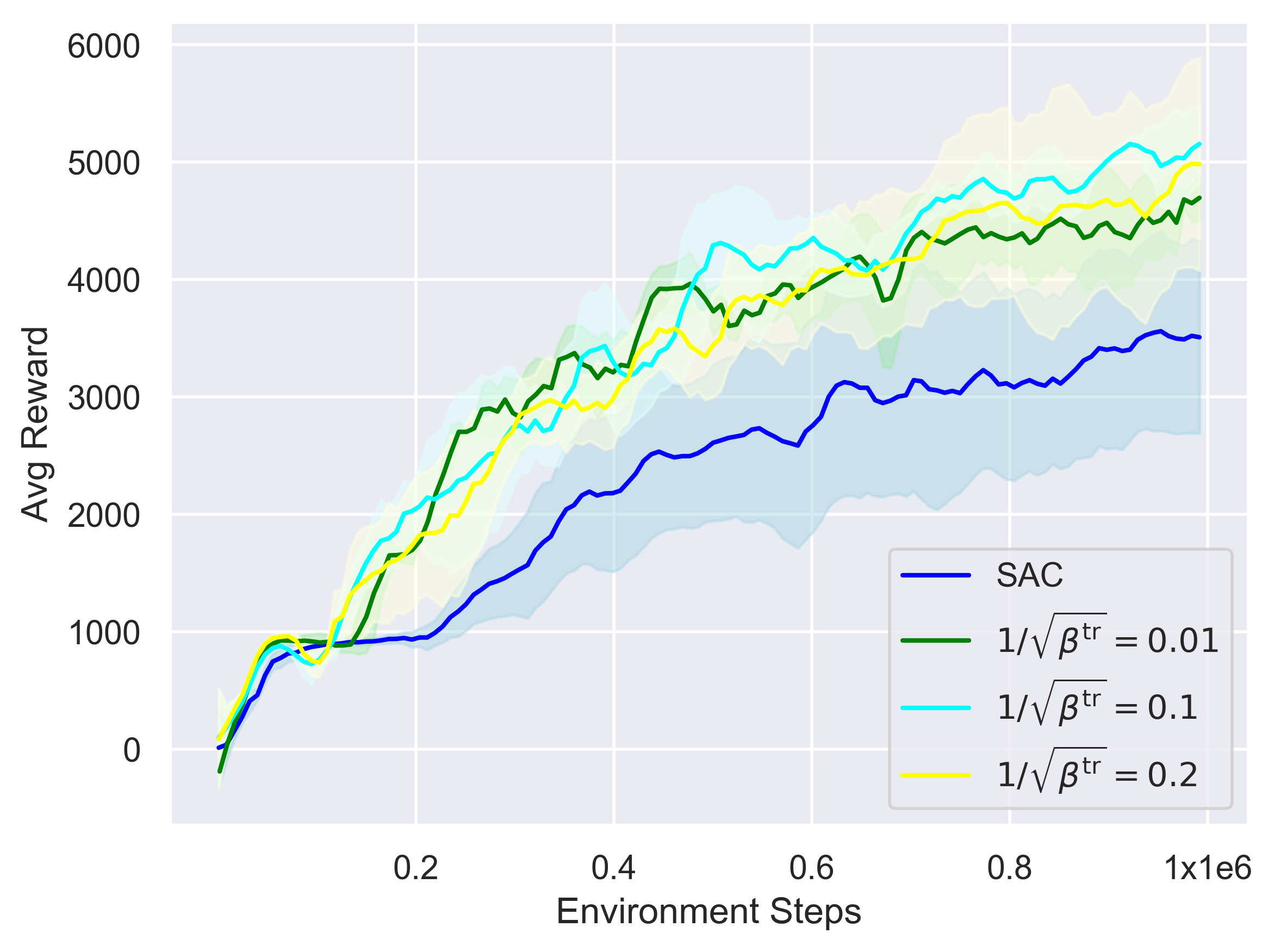}
    \label{fig:sens_2}
}
\subfigure[PyBullet Flagrun Harder]{
    \includegraphics[width=1.5in]{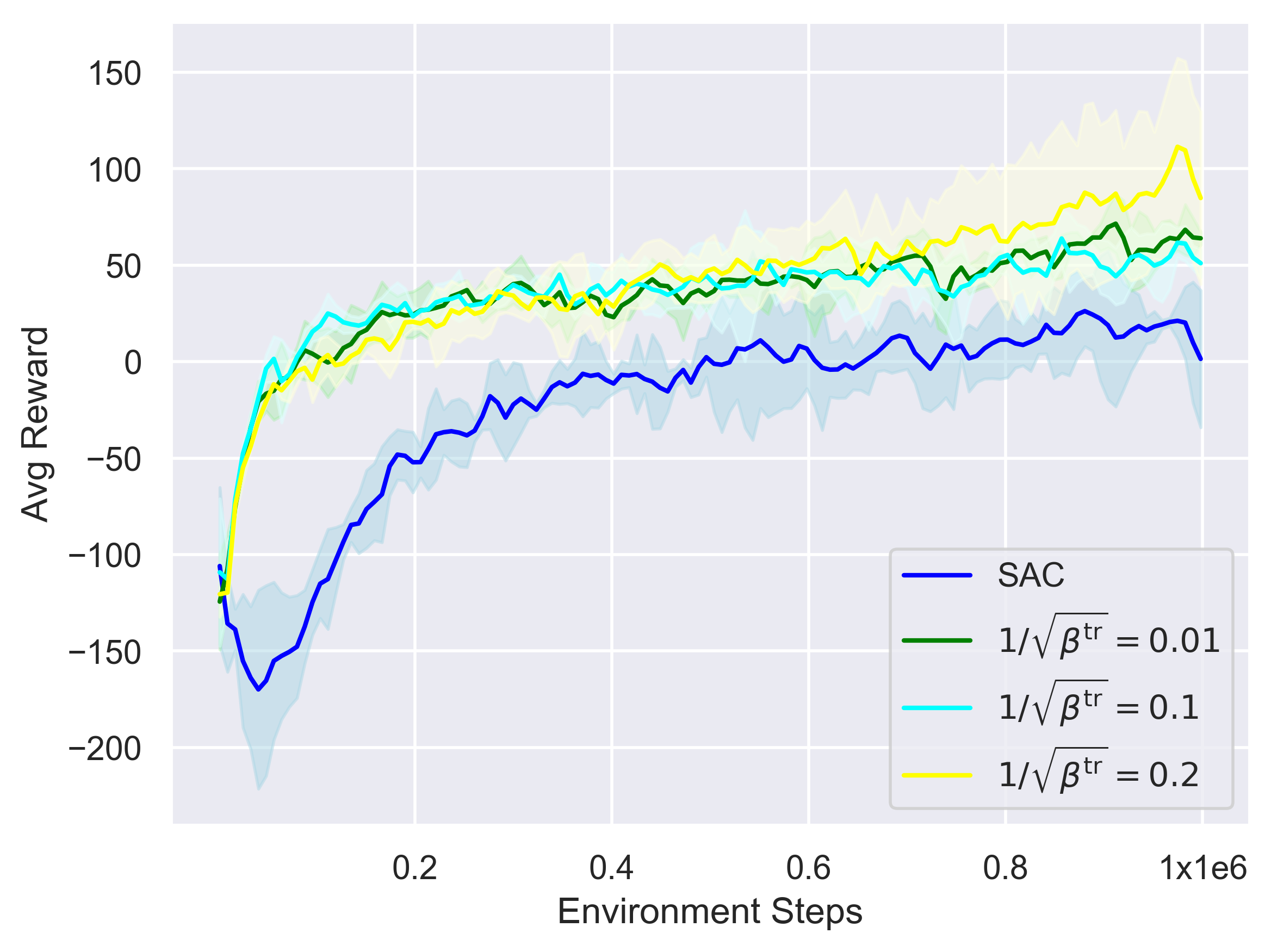}
    \label{fig:sens_3}
}

\subfigure[Hopper-v2]{
    \includegraphics[width=1.5in]{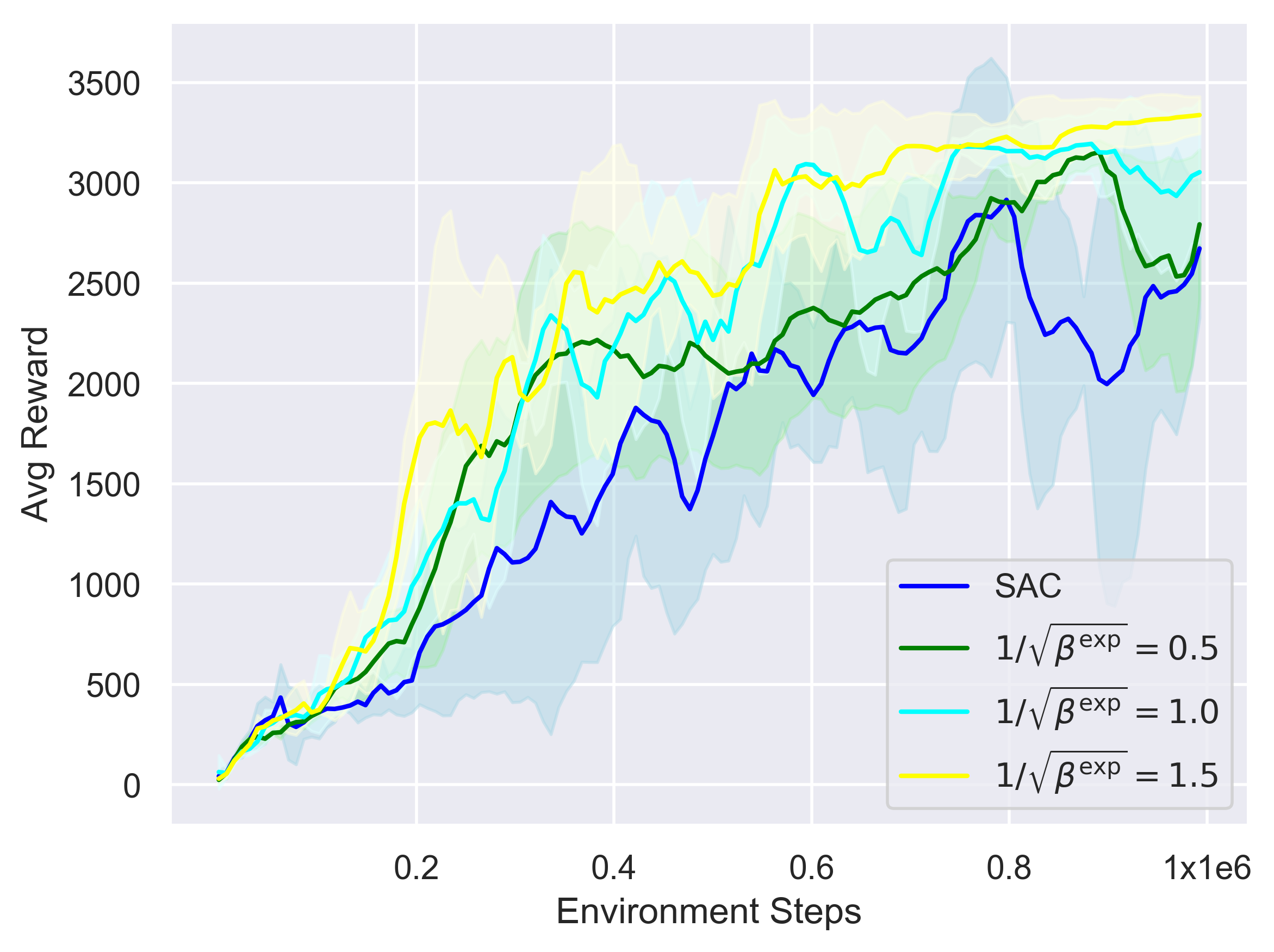}
    \label{fig:sens_4}
}
\subfigure[Ant-v2]{
    \includegraphics[width=1.5in]{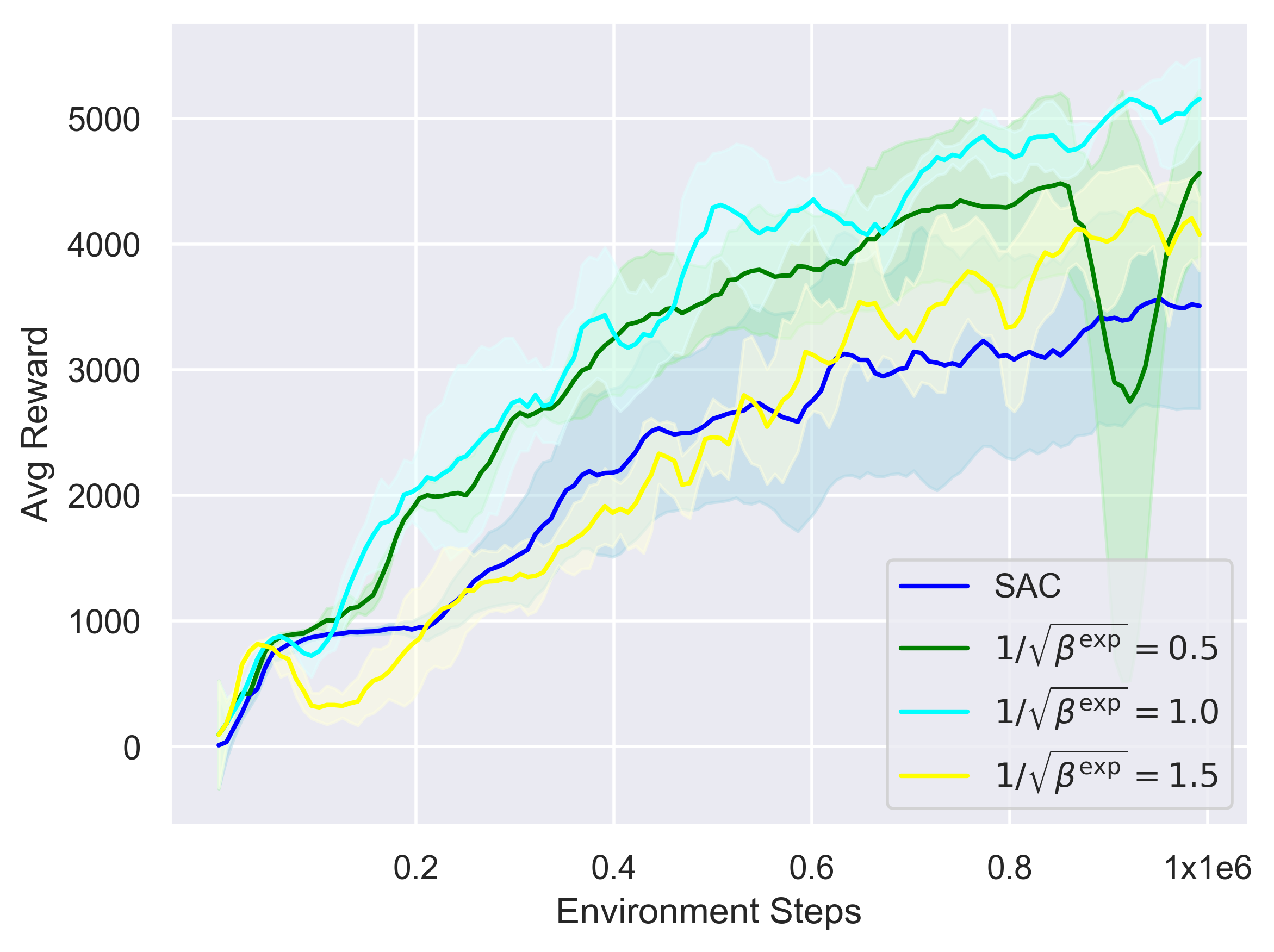}
    \label{fig:sens_5}
}
\subfigure[PyBullet Flagrun Harder]{
    \includegraphics[width=1.5in]{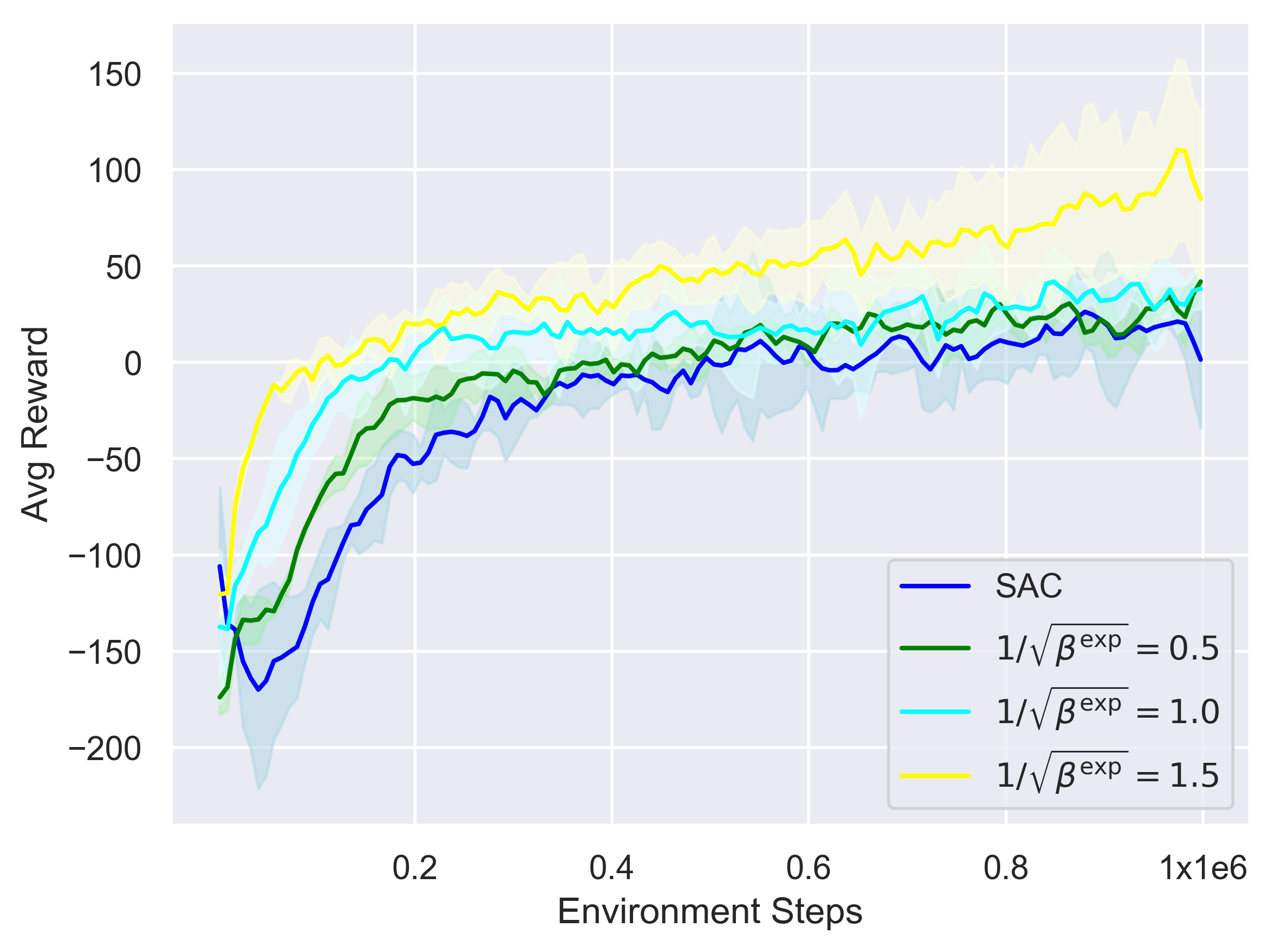}
    \label{fig:sens_6}
}

\subfigure[Hopper-v2]{
    \includegraphics[width=1.5in]{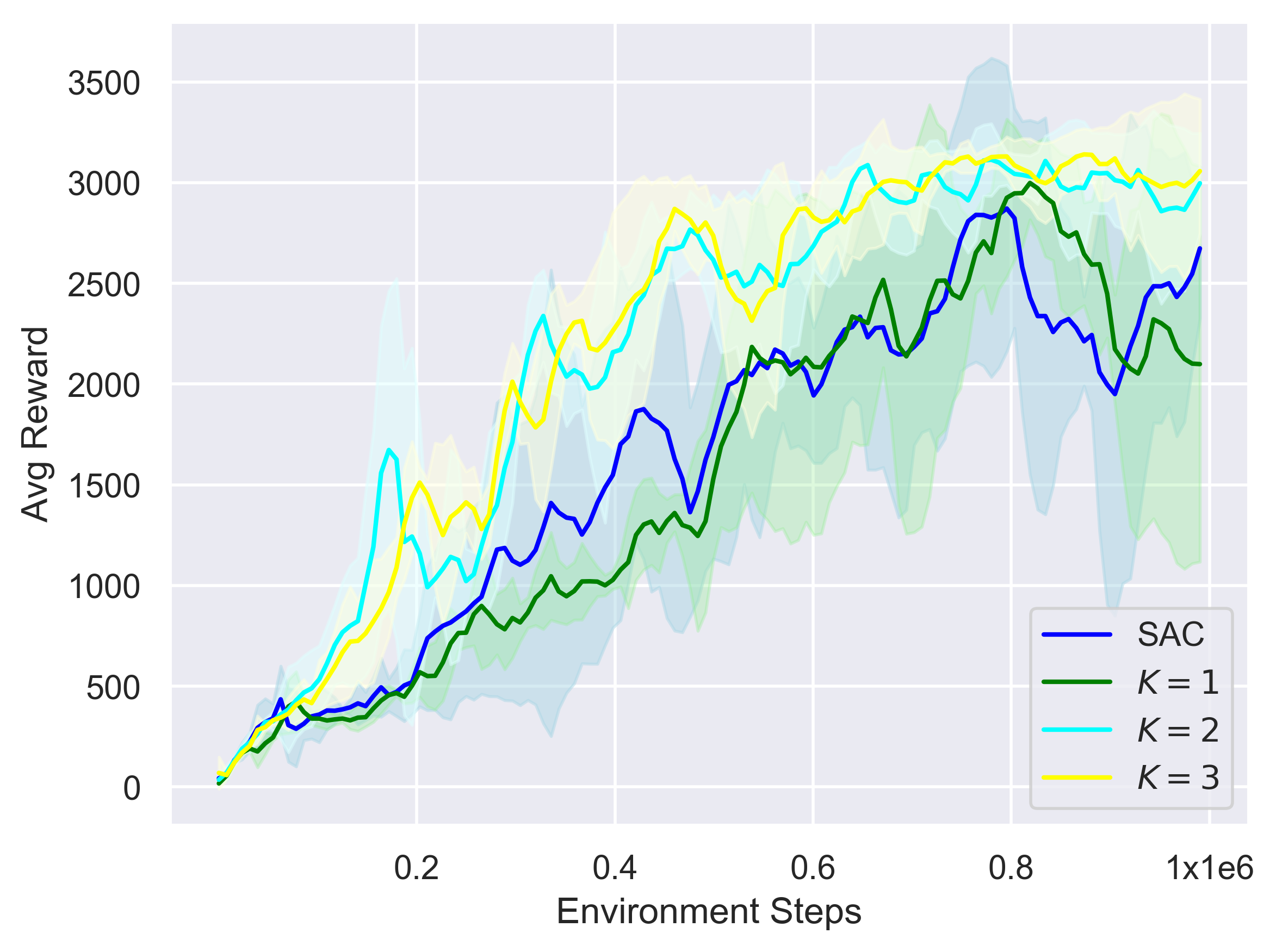}
    \label{fig:sens_7}
}
\subfigure[Ant-v2]{
    \includegraphics[width=1.5in]{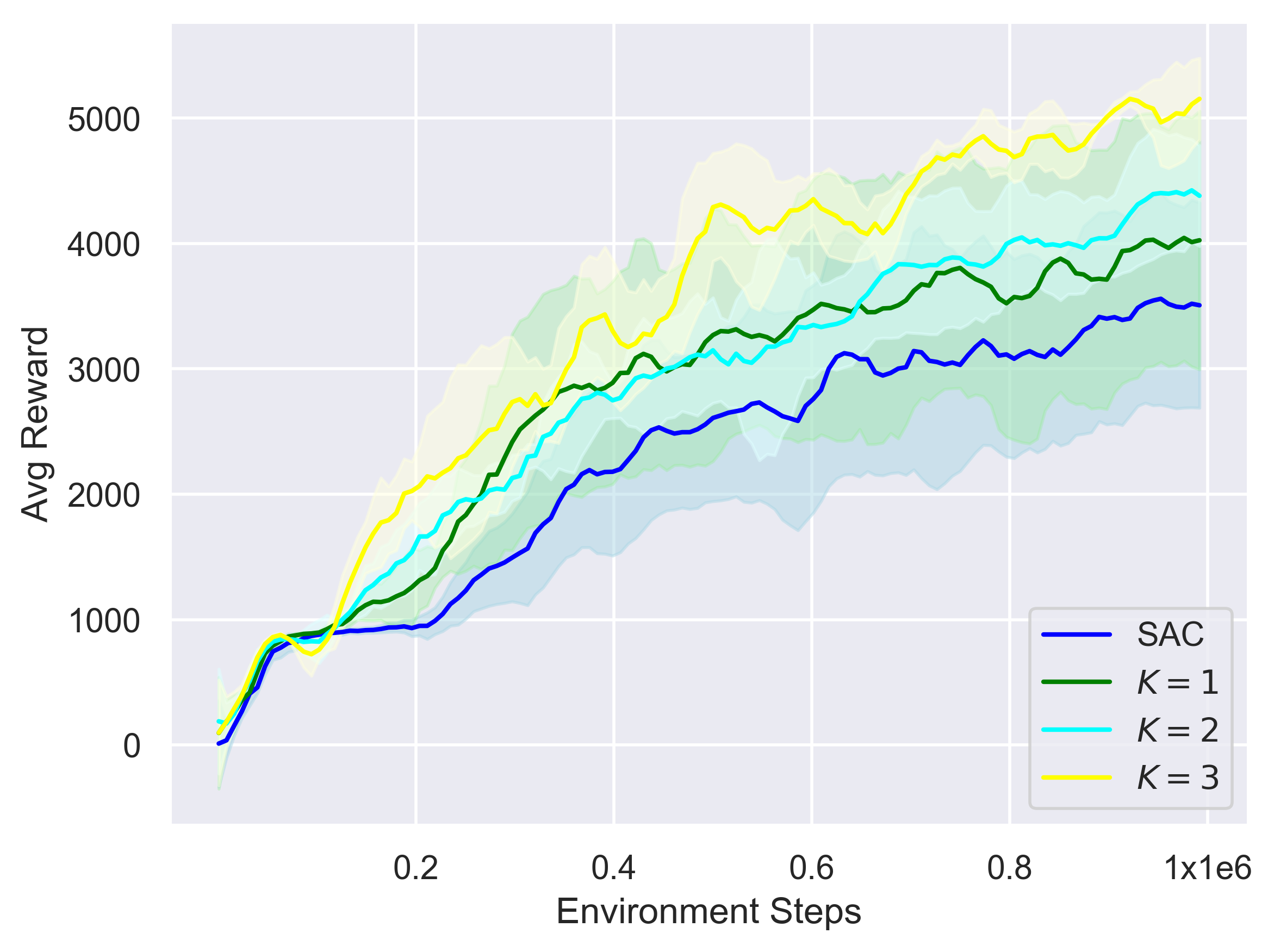}
    \label{fig:sens_8}
}
\subfigure[PyBullet Flagrun Harder]{
    \includegraphics[width=1.5in]{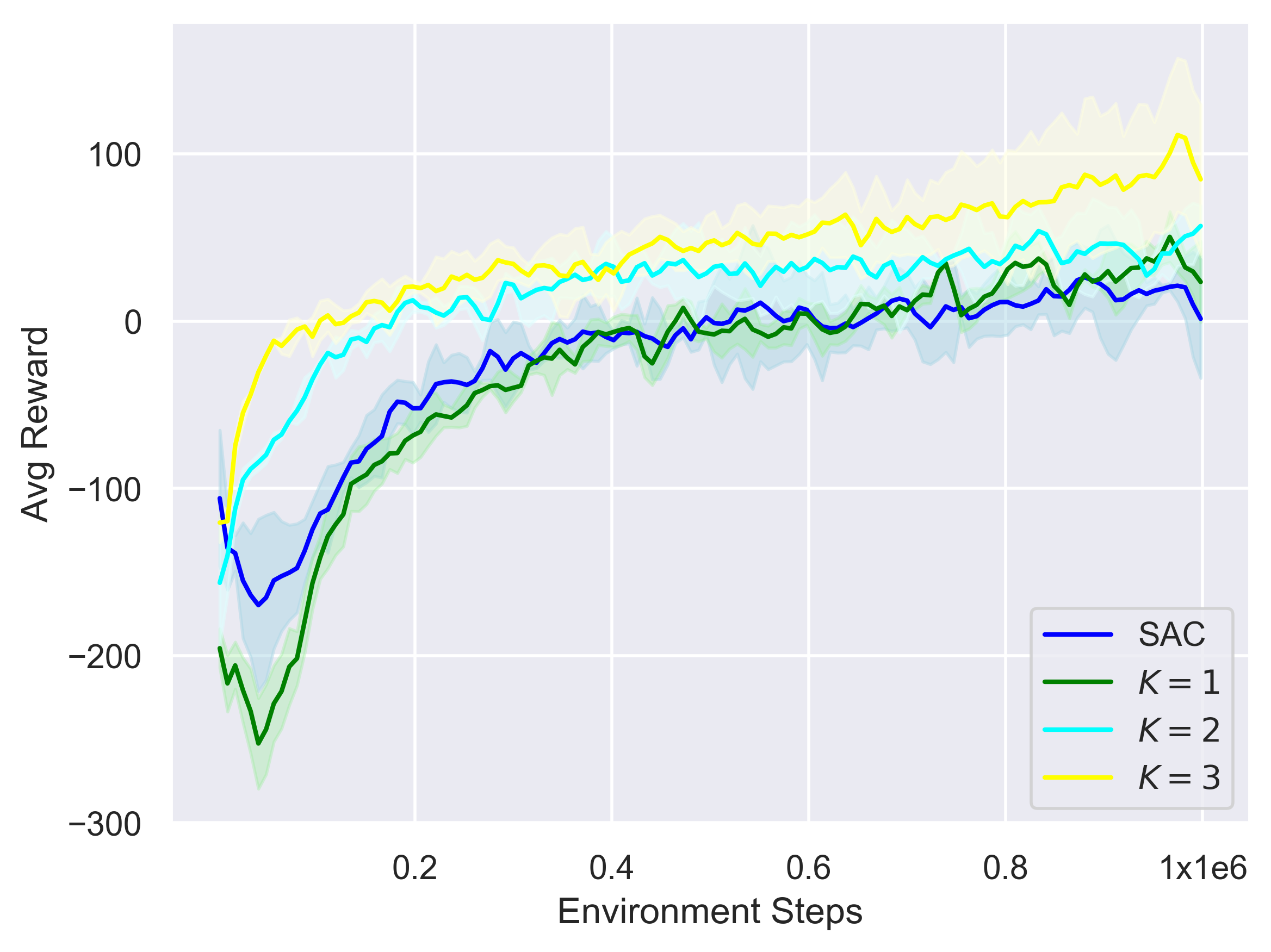}
    \label{fig:sens_9}
}
\caption{Sensitivity analysis on $K, \beta^{\text{exp}}$, and $\beta^{\text{tr}}$.}
\label{fig:sens}
\end{figure*}

\subsection{Analysis}
\label{sec:analysis}
In this section, we conduct ablation study, sensitivity analysis and investigate the shape of the policy distributions evolved by HD.

\subsubsection{Ablation Study}
\label{sec:abl}
In ablation study, we first verify the effect of the proposed leapfrog operator \eqref{leapfrog1} in comparison with the conventional leapfrog in \eqref{leapfrog}. Then we study the differences of the effects of HMC in exploration and policy training, where the policy training (policy optimization) refers to the training step for policy network and neural networks $T_h, \sigma_h$ in leapfrog \eqref{leapfrog1}. 

Specifically, we introduce three baselines adopting different leapfrog operators in exploration and policy training, which are summarized in Table \ref{tab:abl}. Here "Gaussian Policy" means that the same policy network in SAC is directly used without evolving actions by HMC. "Conv. Leapfrog" refers to that actions are evolved by the conventional leapfrog in \eqref{leapfrog}, and "Prop. Leapfrog" denotes that actions are evolved by the proposed leapfrog in \eqref{leapfrog1}. We cannot use the proposed leapfrog only in exploration, since the neural networks $T_h, \sigma_h$ in \eqref{leapfrog1} need to be trained. 

\begin{table}[H]
\centering
\begin{tabular}{p{0.6in}<{\centering}|p{1in}<{\centering}|p{1in}<{\centering}}
\hline
 & Exploration & Policy Training \\\hline
SAC-HPO & Prop. Leapfrog & Prop. Leapfrog \\
Baseline-1 & Conv. Leapfrog & Conv. Leapfrog \\
Baseline-2 & Conv. Leapfrog & Gaussian Policy \\
Baseline-3 & Gaus. Policy & Prop. Leapfrog \\
SAC & Gaussian Policy & Gaussian Policy \\
\hline
\end{tabular}
\caption{Exploration and Training Strategies in Baselines}
\label{tab:abl}
\end{table}

The SAC-HPO and baselines are evaluated over Ant-v2 and HumanoidPyBulletEnv-v0, and learning curves are shown in Figure \ref{fig:ablation}, where we use the same hyper-parameters as Section \ref{sec:mujoco}. It can be seen that in Figure \ref{fig:ablation_1} and \ref{fig:ablation_3}, SAC-HPO outperforms the baseline-1 in terms of both performance and learning stability, showing the advantage of the proposed leapfrog operator over the conventional counterpart. And this advantage can also be observed in other environments.

In Figure \ref{fig:ablation_2} and \ref{fig:ablation_4}, the baseline-1 outperforms both baseline-2 and baseline-3, showing the effects of leapfrog in both exploration and policy optimization. Further, we can see that the improvement of baseline-2 over SAC is higher than that of baseline-3 over SAC, meaning that using HMC in exploration can yield more performance improvement than that in policy training. %We know that to improve policy learning performance, making the exploration directionally informed is more important than reducing amortization gap.

\subsubsection{Sensitivity Analysis}
\label{sec:sens}

\begin{figure*}
\centering
\subfigure[Hopper-v2]{
    \includegraphics[width=1.5in]{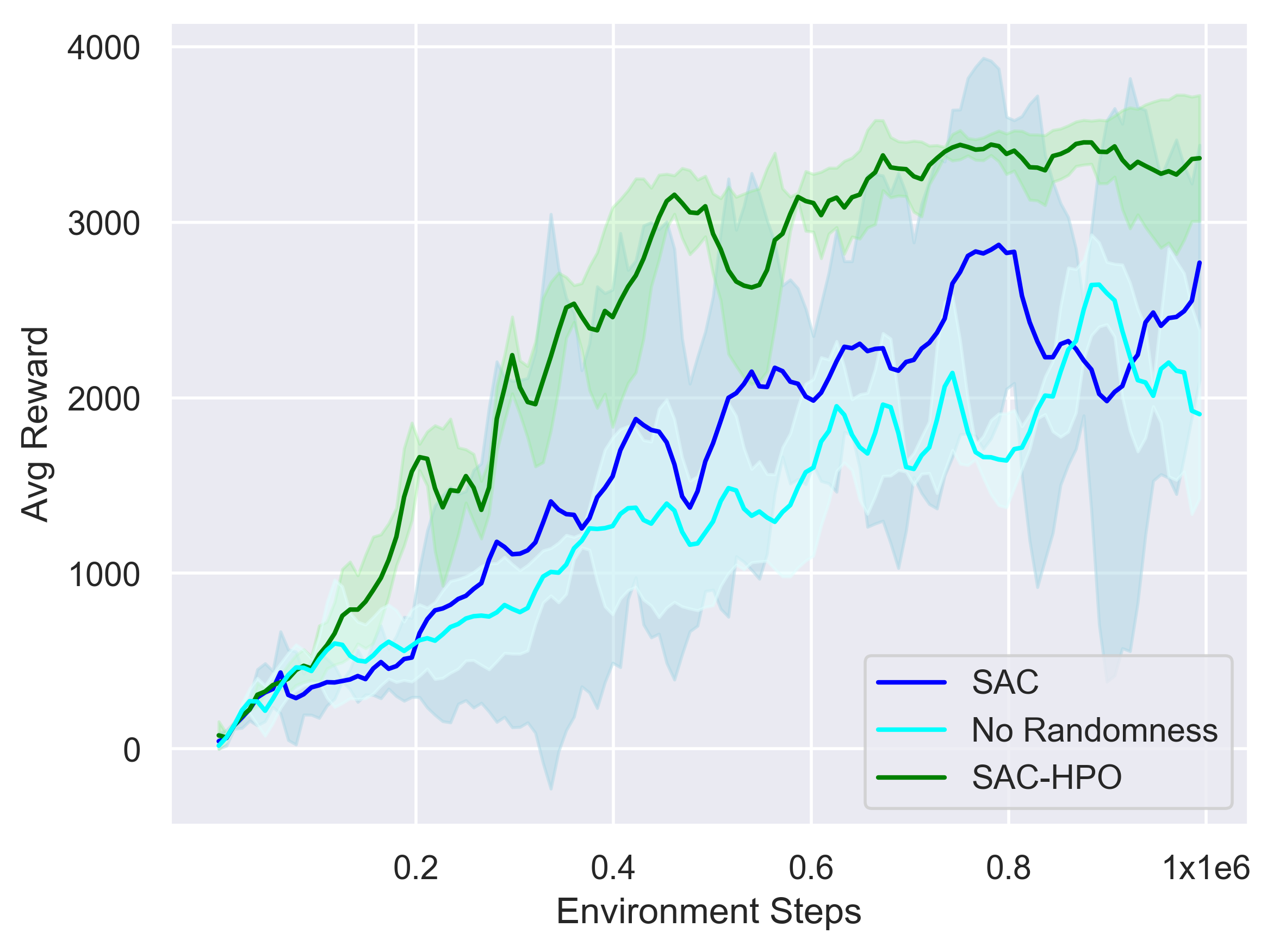}
    \label{fig:nonoise_1}
}
\subfigure[Ant-v2]{
    \includegraphics[width=1.5in]{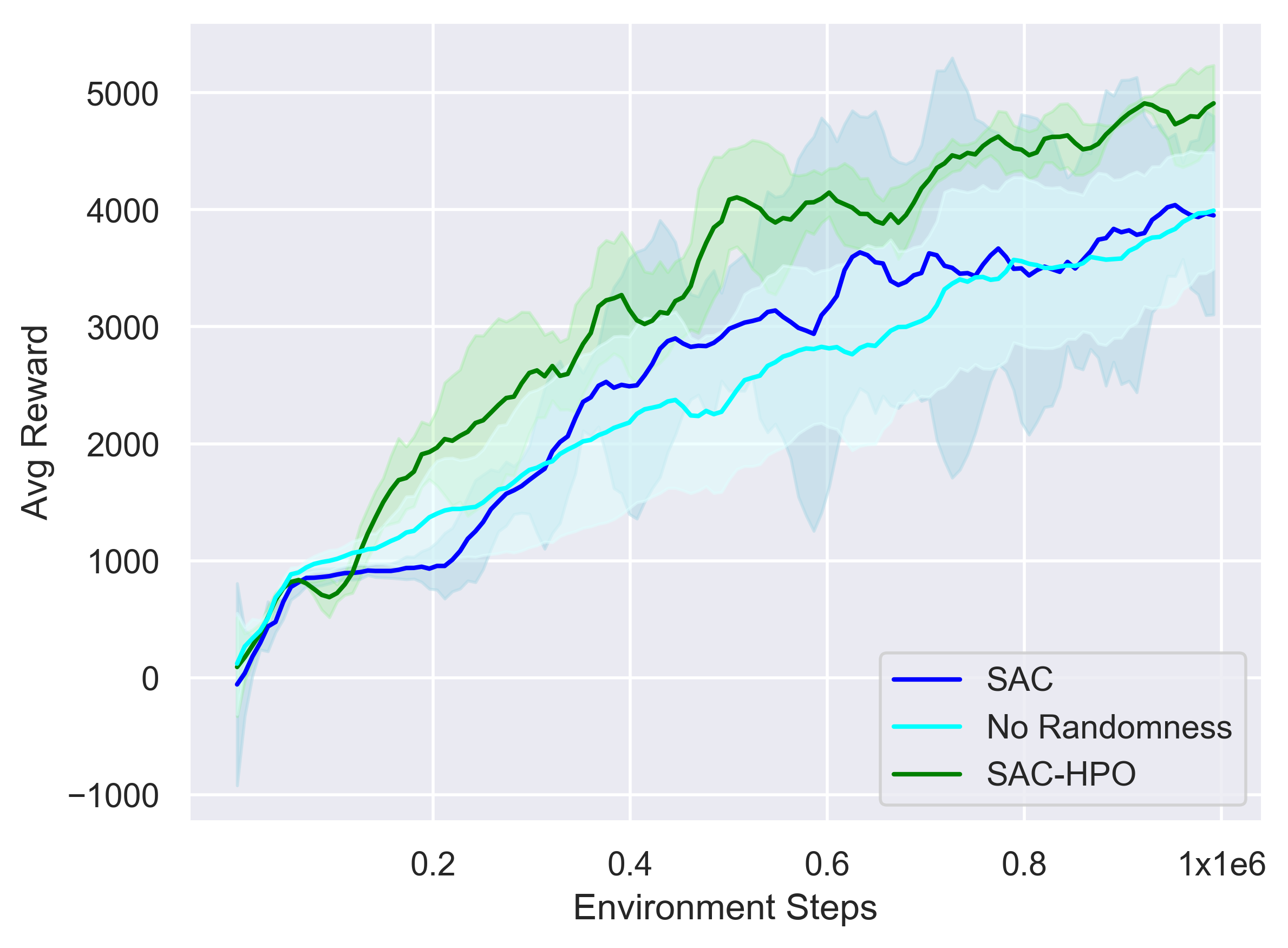}
    \label{fig:nonoise_2}
}
\subfigure[PyBullet Flagrun Harder]{
    \includegraphics[width=1.5in]{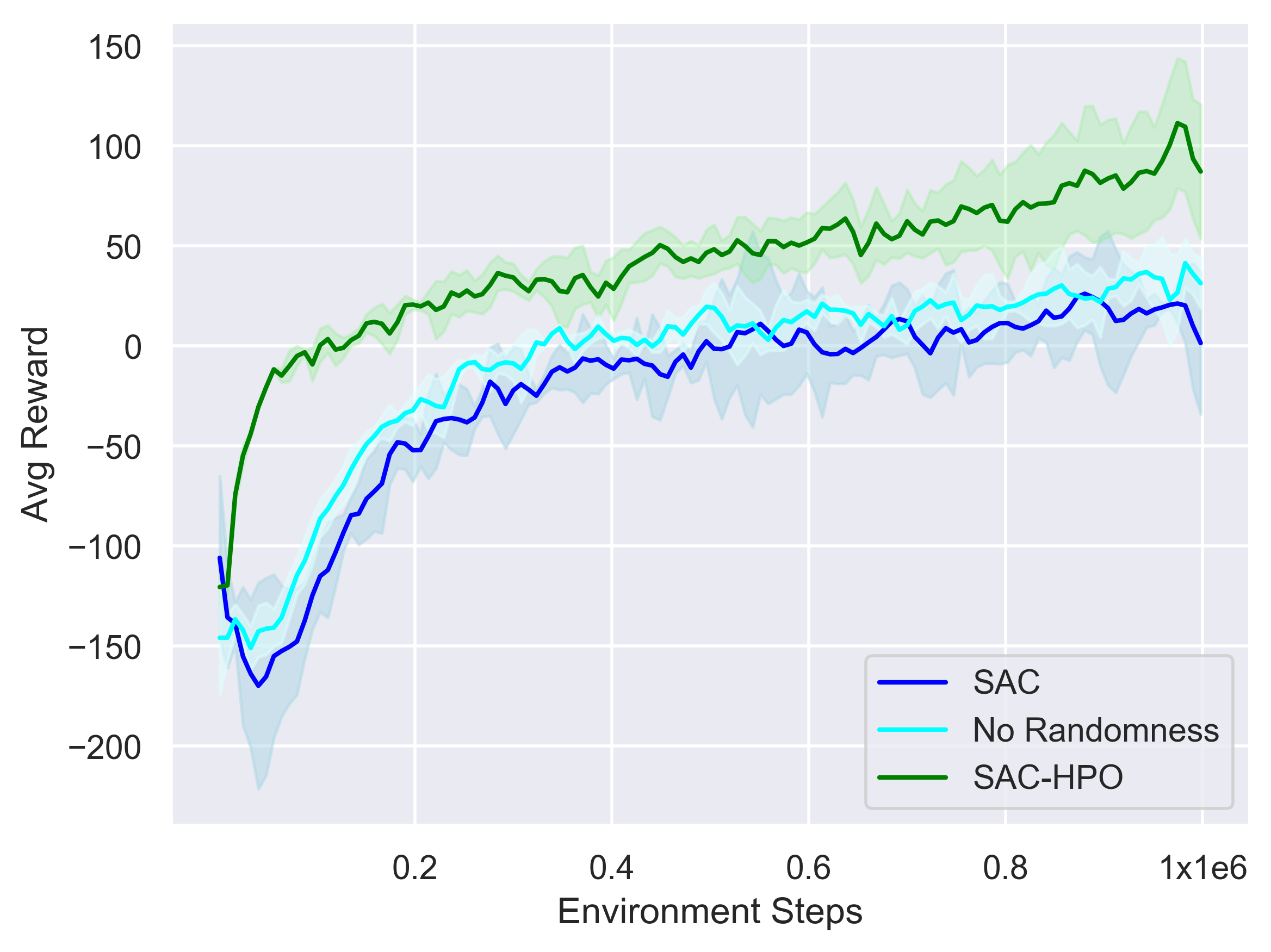}
    \label{fig:nonoise_3}
}
\caption{Performance comparison on the randomness of momentum vector. "No Randomness" refers to SAC-HPO without using random momentum vectors which is to set $\rho=0$.}
\label{fig:nonoise}
\end{figure*}

\begin{figure*}[ht]
\centering
\subfigure[270K Step]{
    \includegraphics[width=1.6in]{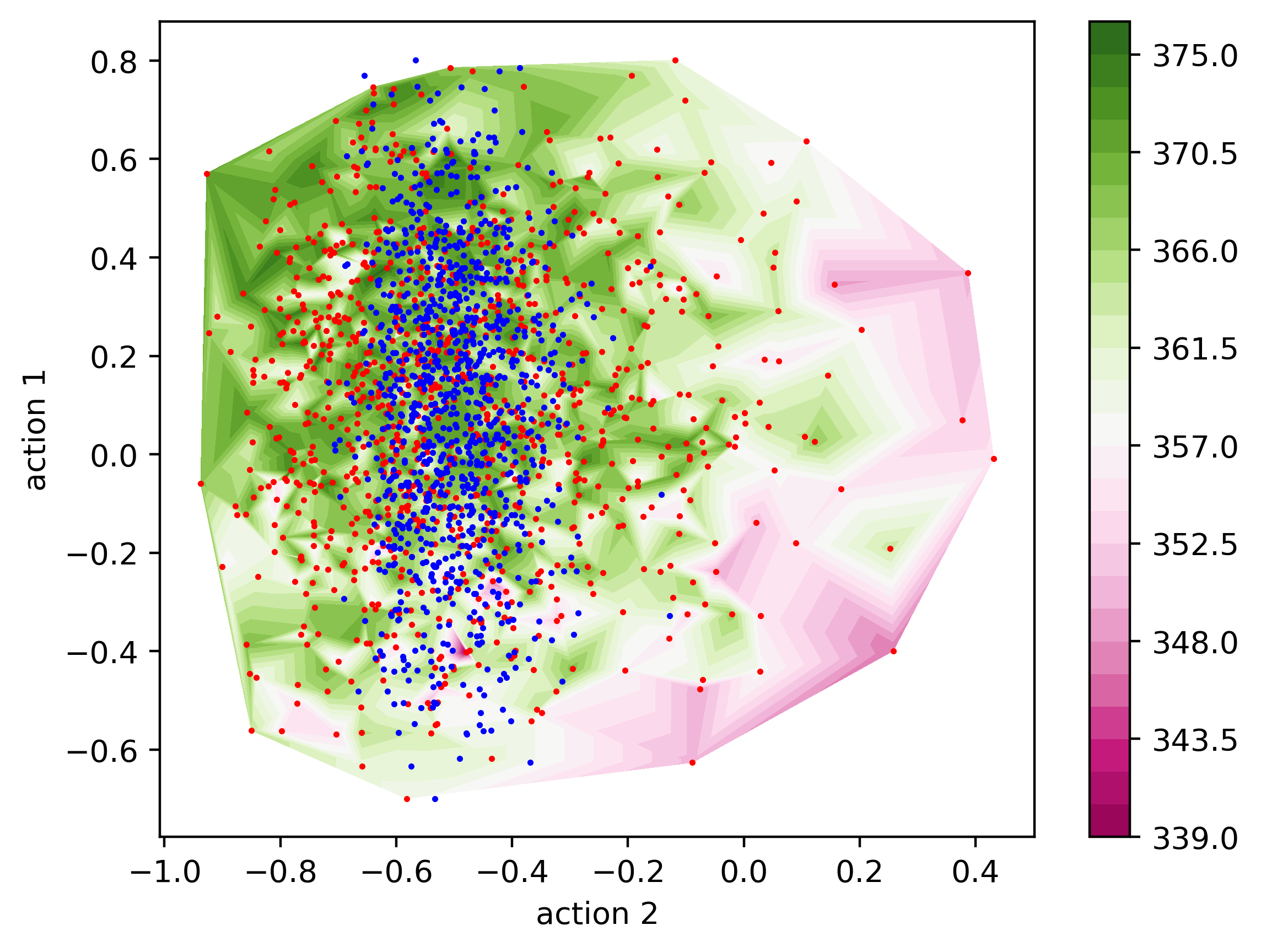}
    \label{fig:shape_1}
}
\subfigure[580K Step]{
    \includegraphics[width=1.6in]{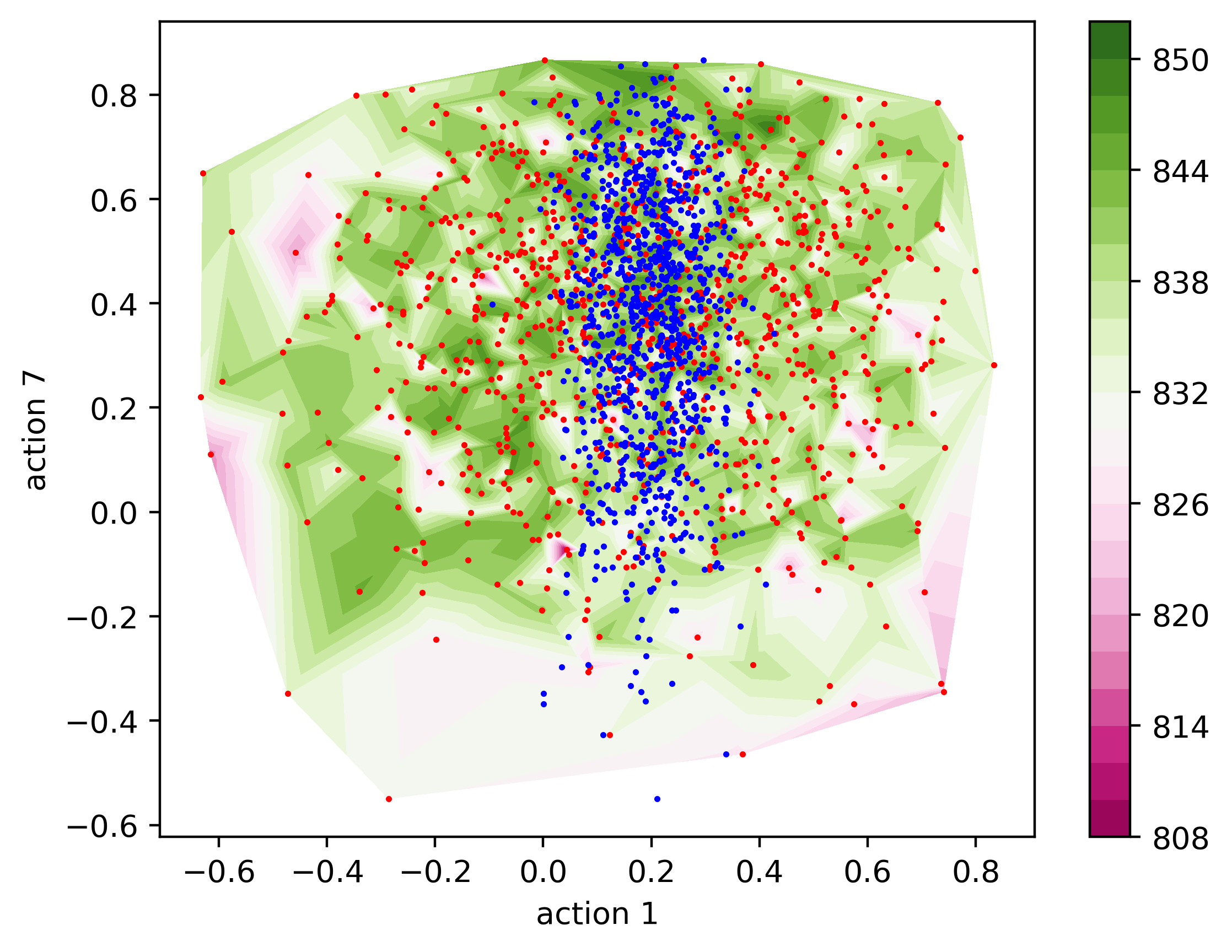}
    \label{fig:shape_2}
}
\subfigure[880K Step]{
    \includegraphics[width=1.6in]{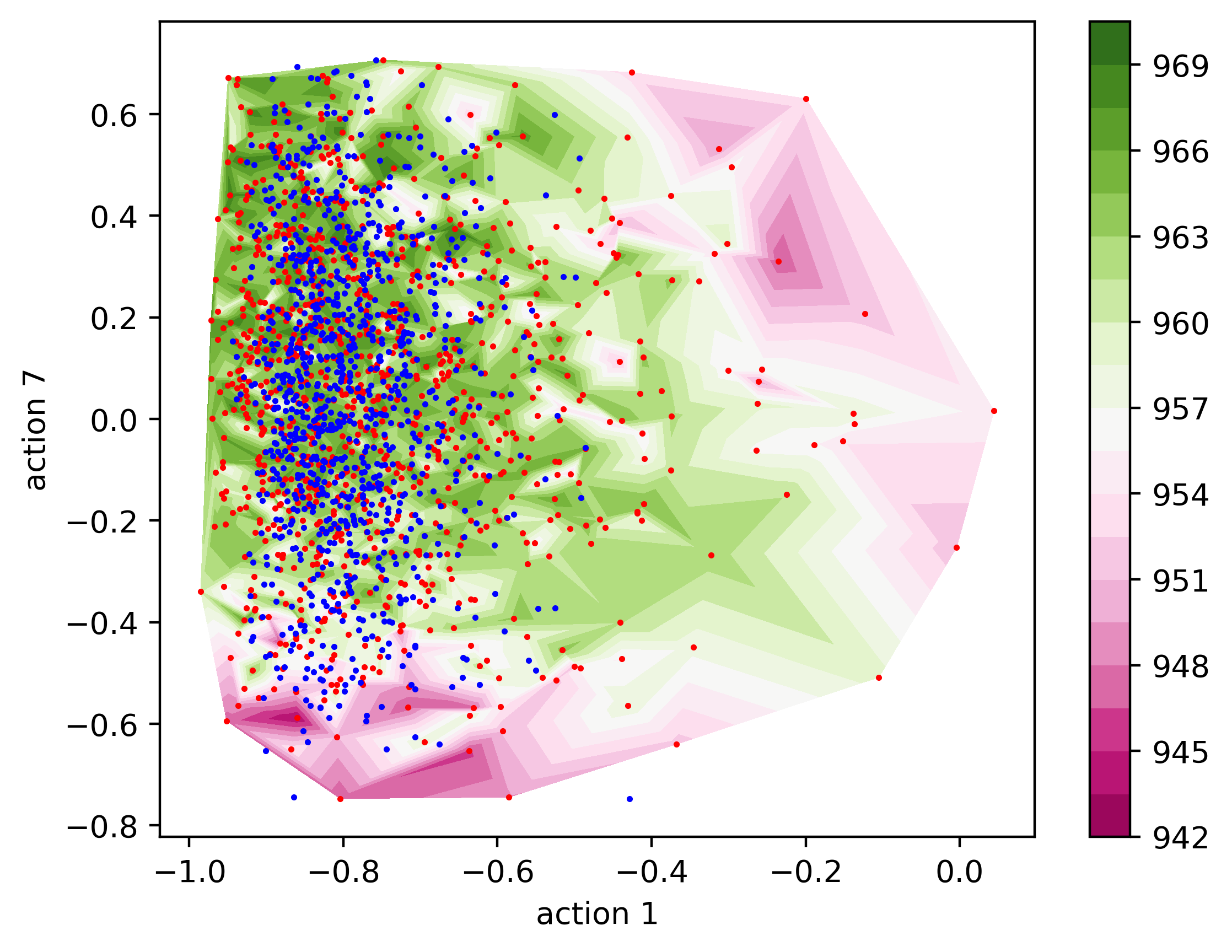}
    \label{fig:shape_3}
}
\caption{The shape of policy distribution in Ant-v2. The dimensions of action are shown as x and y labels. The color bar is for Q values. $\beta_0^{\text{exp}}=1$ for every step.}
\label{fig:shape}
\end{figure*}

Here we conducted sensitivity analysis on three important hyper-parameters. The first is the number of leapfrog steps in simulating HD, i.e., $K=1,2,3$, shown in Figures \ref{fig:sens_7}, \ref{fig:sens_8} and \ref{fig:sens_9}. We can see that in Hopper-v2, the cases of $K=2$ and $K=3$ perform similar, but much better than that of $K=1$, showing that it is not meaningful to have larger $K$. The second parameter analyzed here is the variance of momentum vector $\rho$ in exploration, i.e., $1/\sqrt{\beta^{\text{exp}}_0}=0.5, 1.0, 1.5$, shown in Figure \ref{fig:sens_4}, \ref{fig:sens_5} and \ref{fig:sens_6}. We can see that the performance is sensitive to the choice of $\beta^{\text{exp}}$, showing the variance of $\rho$ in exploration is important to the learning performance. Although more randomness in exploration can encourage the agent to explore more unseen states, not the highest choice of $\beta^{\text{exp}}$ leads to best performance, such as Figure \ref{fig:sens_5}. That is because too much variance of $\rho$ may make the sampled state-action pairs deviate too much from the optimal trajectory during the exploration. In Figure \ref{fig:sens_1}, \ref{fig:sens_2} and \ref{fig:sens_3}, we analyze three different values for the variances for momentum vector $\rho$ in policy training, i.e., $1/\sqrt{\beta_0^{\text{tr}}}=0.01, 0.1, 0.2$. However, we can see that the performance is not sensitive to $\beta_0^{\text{tr}}$. According to more evaluations, larger $\beta_0^{\text{tr}}$ cannot lead to better performance. For example when $\beta_0^{\text{tr}}=1$, the performance degrades significantly.

\subsubsection{Effect of Random Momentum Vector}
\label{sec:nonoise}
In Figure \ref{fig:nonoise}, we study the effect of the randomness of momentum vector $\rho$ on the performance improvement. It is similar as the comparison with iterative amortized policy optimization (IAPO) \cite{marino2020iterative}. The baseline here is the SAC-HPO without random momentum vector, where we set $\rho=0$ in both exploration and policy training. We can see that if no randomness in momentum vectors, the performance of SAC-HPO degrades significantly. Previous work such as IAPO \cite{marino2020iterative} directly uses gradients to update the actions sampled from the base policy, without using any extra random variables, which is same as the baseline here. However, their performance is not good in high-dimensional environments \cite{marino2020iterative}. The comparison in Figure \ref{fig:nonoise} can explain the reason and show the advantage of our method. And similar advantage of our method can be observed in other environments as well.

\subsubsection{Visualization of Policy Distribution}
\label{sec:shape}
In Figure \ref{fig:shape}, we visualize the action distributions of Hamiltonian policy (actions evolved by HMC) at different environmental steps, where x and y axes represent two different action dimensions. Specifically, in Figure \ref{fig:shape}, the red dots represent 1000 actions sampled from the policy distribution evolved by leapfrog steps \eqref{leapfrog1}. For comparison, the blue dots represent 1000 actions sampled from the base policy network $\pi_{\theta}$ directly, which are Gaussian and are not evolved by HMC. The contour of Q values is shown as background for reference, which is drawn by triangular interpolation method. In Figure \ref{fig:shape}, comparing red and blue dots, we can see that HMC can evolve actions sampled from the base policy more towards regions with higher Q values, making sampled actions more directionally informed and hence improving exploration efficiency. We can also observe that policy distribution evolved by leapfrog operators can be highly non-Gaussian and have larger variance with much broader effective support. Besides, there are still some actions evolved to regions with similar or lower Q values, so that a reasonable trade-off of exploration and exploitation can be reached. That is why the exploration of RL agent can be boosted by Hamiltonian policy and the learning performance can be improved. 
%In safe RL problems, the Hamiltonian policy cannot only improve the achieved return but also reduce the number of safety constraint violations. 

\section{Conclusion}
In this work, we propose to integrate policy optimization with HMC, evolving actions from the base policy network by Hamiltonian dynamics simulated by leapfrog steps. In order to adapt to the changes of Q functions which define the target posterior, we propose a new leapfrog operator which generalizes HMC via gated neural networks. The proposed method can improve the efficiency of policy optimization and make the exploration more directionally informed. In empirical experiments, the proposed method can outperform baselines in terms of both convergence rate and performance. In safe RL problems, the Hamiltonian policy cannot only improve the achieved return but also reduce the number of safety constraint violations.